\documentclass[10pt,twocolumn,letterpaper]{article}

\usepackage{wacv}
\usepackage{times}
\usepackage{epsfig}
\usepackage{graphicx}
\usepackage{amsmath}
\usepackage{amssymb}
\usepackage{siunitx}
\usepackage{xspace}
\usepackage{color,xcolor}
\usepackage{booktabs}       %
\usepackage{microtype}      %
\usepackage{enumitem}
\usepackage{algorithm}
\usepackage{algorithmic}
\usepackage{subcaption}
\usepackage{caption}
\usepackage[xindy,acronym,toc,smallcaps,nomain]{glossaries}
\usepackage[accsupp]{axessibility}

\def\wacvPaperID{802} %

\wacvfinalcopy %

\ifwacvfinal
\def\assignedStartPage{9876} %
\fi

\ifwacvfinal
\usepackage[breaklinks=true,bookmarks=false]{hyperref}
\else
\usepackage[pagebackref=true,breaklinks=true,colorlinks,bookmarks=false]{hyperref}
\fi

\ifwacvfinal
\else
\fi

\newacronym{ANASOD}{anasod}{Approximate Neural Architecture Search with Operation Distribution}
\newacronym{NAS}{nas}{Neural Architecture Search}
\newacronym{DARTS}{darts}{Differentiable Neural Architecture Search}
\newacronym{DNAS}{dnas}{Differentiable Neural Architecture Search}
\newacronym{NAS-Bench-101}{nb101}{NAS-Bench-101}
\newacronym{NAS-Bench-201}{nb201}{NAS-Bench-201}
\newacronym{NAS-Bench-301}{nb301}{NAS-Bench-301}
\newacronym{CIFAR-10}{cifar-10}{dummy}
\newacronym{CIFAR-100}{cifar-100}{dummy}
\newacronym{LS}{ls}{Local Search}
\newacronym{RS}{rs}{Random Search}
\newacronym{SMBO}{smbo}{Sequential Model-based Optimisation}
\newacronym{BO}{bo}{Bayessian Optimisation}
\newacronym{GP}{gp}{Gaussian Process}
\newacronym{NEP}{nep}{Neural Ensemble Predictor}
\newacronym{SD}{sd}{standard deviation}

\glsunset{NAS}
\glsunset{ANASOD}
\glsunset{DARTS}
\glsunset{CIFAR-10}
\glsunset{CIFAR-100}

\newcommand{\our}{\gls{ANASOD}\xspace}

\begin{document}

\title{Approximate Neural Architecture Search via Operation Distribution Learning}

\author{Xingchen Wan$^{1,2}$\quad Binxin Ru$^{1,2}$\quad Pedro M. Esperança$^{1}$ \quad Fabio M. Carlucci$^1$\\
\small{$^1$Huawei Noah's Ark Lab, London, UK \qquad $^2$Machine Learning Research Group, University of Oxford, Oxford, UK}\\
{\tt\small \{xwan,robin\}@robots.ox.ac.uk pedro.esperanca@huawei.com fabiom.carlucci@gmail.com}
}

\maketitle

\begin{abstract}

The standard paradigm in Neural Architecture Search (\gls{NAS}) is to search for a fully deterministic architecture with specific operations and connections.
In this work, we instead propose to search for the optimal \textbf{operation distribution}, thus providing a stochastic and approximate solution, which can be used to sample architectures of arbitrary length. We propose and show, that given an architectural cell, its performance largely depends on the ratio of used operations, rather than any specific connection pattern in typical search spaces; that is, small changes in the ordering of the operations are often irrelevant. This intuition is orthogonal to any specific search strategy and can be applied to a diverse set of \gls{NAS} algorithms. 
Through extensive validation on 4 data-sets and 4 \gls{NAS} techniques (Bayesian optimisation, differentiable search, local search and random search), we show that the operation distribution (1) holds enough discriminating power to reliably identify a solution and (2) is significantly easier to optimise than traditional encodings, leading to large speed-ups at little to no cost in performance. 
Indeed, this simple intuition significantly reduces the cost of current approaches and potentially enable \gls{NAS} to be used in a broader range of applications. 

\end{abstract}

\section{Introduction}

Neural architecture search (\gls{NAS}) is an extremely challenging problem and in the last few years, a huge number of solutions have been proposed in the literature~\cite{elsken2018neural,NEURIPS2020_NAGO,Liu2019_DARTS,ZophLe17_NAS,Pham2018_ENAS}. From genetic algorithms, to Bayesian optimisation, many approaches have been evaluated, each presenting different strengths and weaknesses, with no clear winner overall.
At its core, \gls{NAS} is a combinatorial problem for which no algorithm can guarantee to find the global optimum in polynomial time. For example, the commonly used \gls{DARTS}~\cite{Liu2019_DARTS} search space, even with simplifying constraints, contains $8^{14} \approx 4.3 \times 10^{12}$ architectures, an unreasonably large number to explore. 
Given the sheer size of the search space, it is highly unlikely that \gls{NAS} algorithms are able to reliably find the global optimum and in fact recent research has shown that local search algorithms can perform comparably to the state of the art \cite{white2020local,ottelander2020local}. Furthermore, thanks to a number of benchmarks, we know that many architectures perform similarly one another \cite{ying2019bench,dong2019NASbench201,siems2020bench,Yang2020NASEFH}.

\begin{figure}
    \centering
    \includegraphics[width=0.45\textwidth]{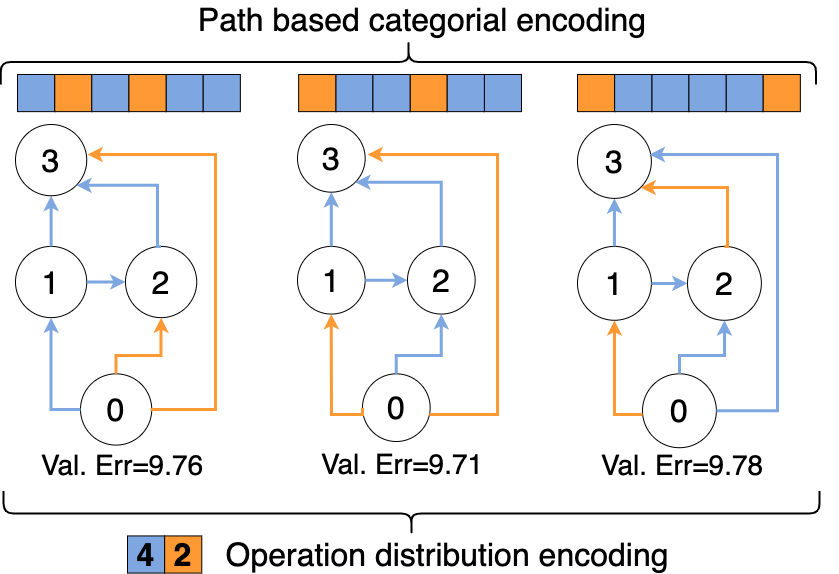}
    \caption{Three architecture cells from \gls{NAS-Bench-201} \cite{dong2019NASbench201} giving very similar validation performance. Each edge represents an operation: either \textcolor{orange}{\texttt{conv1$\times$1}} or \textcolor{blue}{\texttt{conv3$\times$3}}. From the viewpoint of existing encodings, they represent three distinct architectures which all need to be evaluated. The \our framework assumes that they all belong to the same operation distribution ($4\times$\textcolor{blue}{\texttt{conv3$\times$3}} and $2\times$\textcolor{orange}{\texttt{conv1$\times$1}} in 6 operations) and does not repeatedly re-evaluate each one.}
    \label{fig:example_archs}
\end{figure}

\begin{figure*}[t]
    \centering
    \begin{subfigure}{0.24\linewidth}
        \includegraphics[trim=0cm 0cm 0cm  0cm, clip, width=1.0\linewidth]{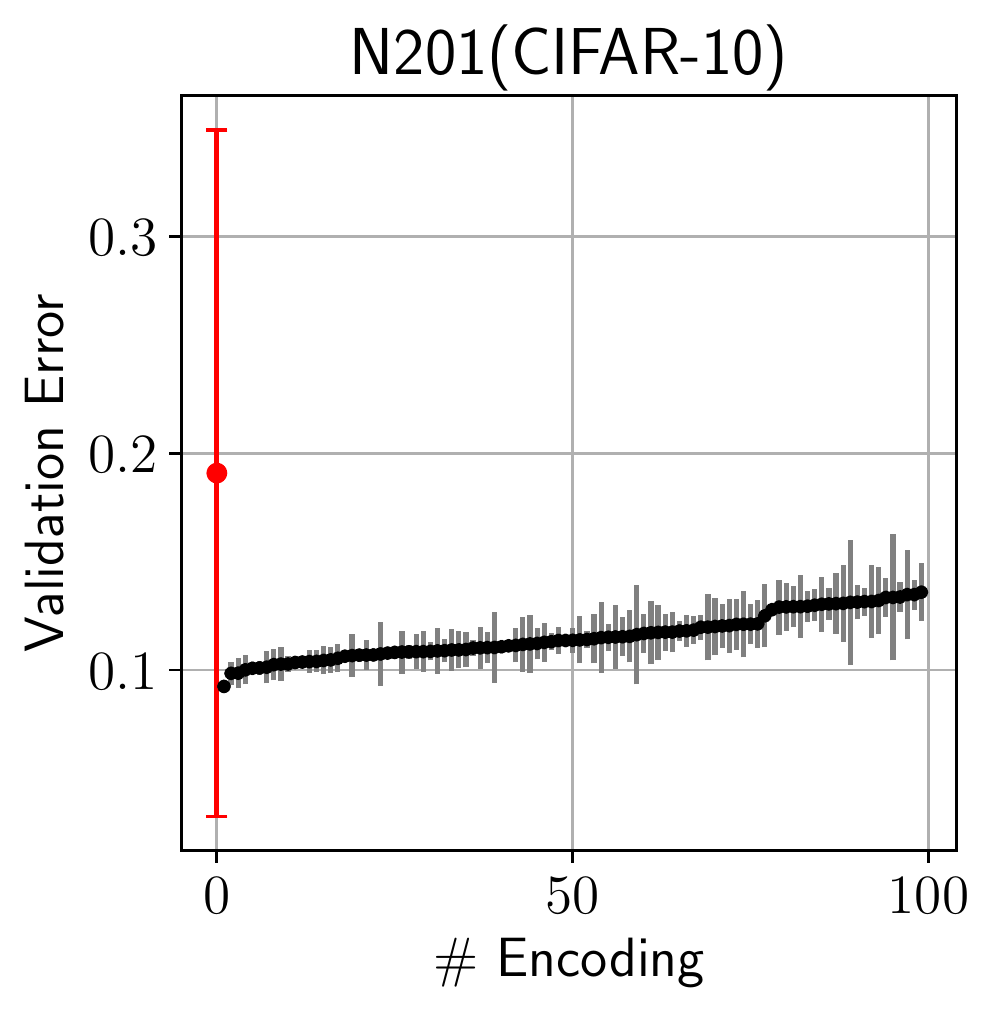}
    \end{subfigure}
    \begin{subfigure}{0.24\linewidth}
        \includegraphics[trim=0cm 0cm 0cm  0cm, clip, width=1.0\linewidth]{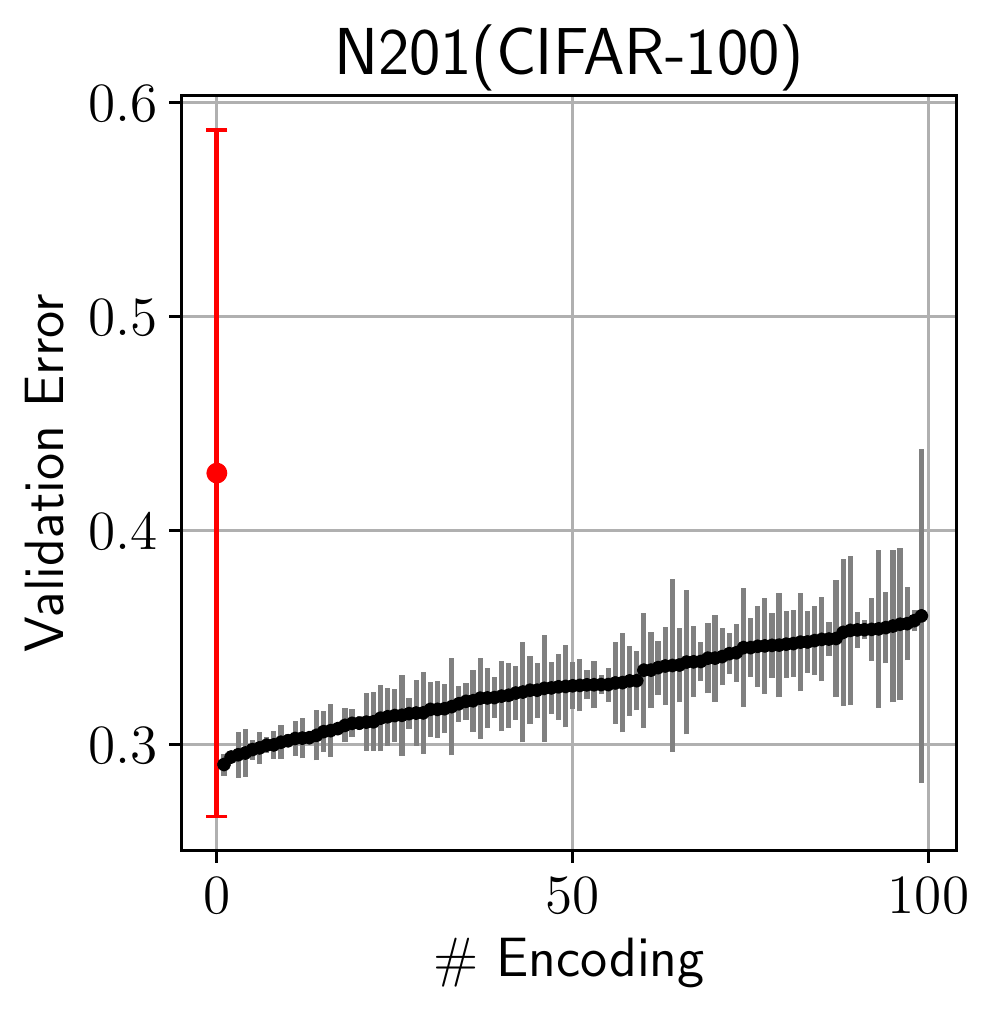}
    \end{subfigure}
        \begin{subfigure}{0.24\linewidth}
        \includegraphics[trim=0cm 0cm 0cm  0cm, clip, width=1.0\linewidth]{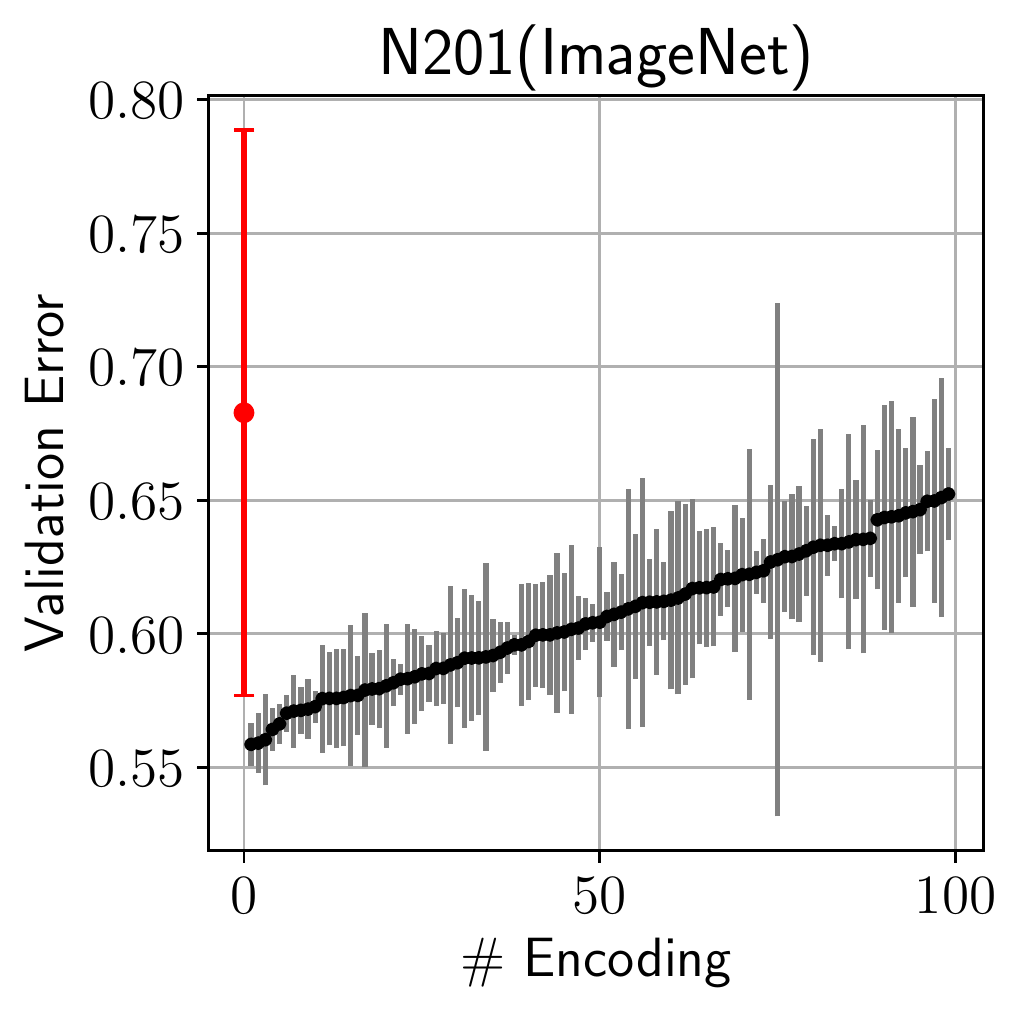}
    \end{subfigure}
        \begin{subfigure}{0.24\linewidth}
        \includegraphics[trim=0cm 0cm 0cm  0cm, clip, width=1.0\linewidth]{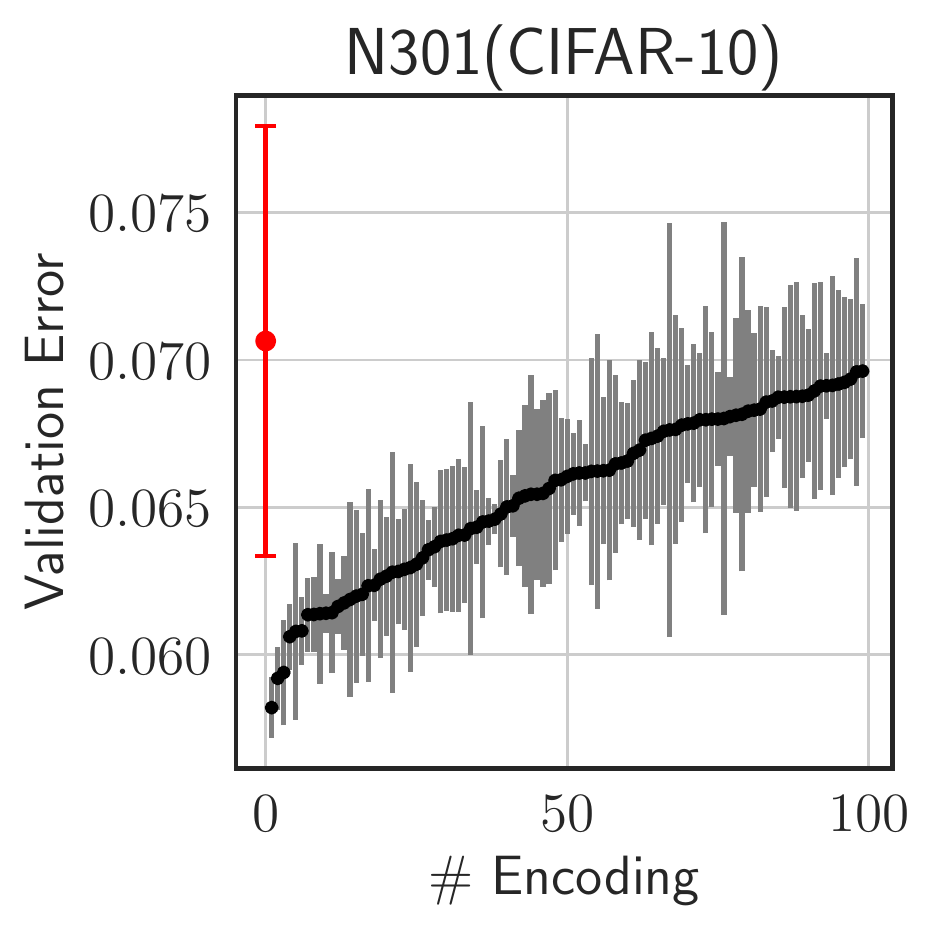}
    \end{subfigure}
    \vspace{-2mm}
    \caption{To what extent does the \our encoding determine performance? We randomly draw $200$ \our encodings in 4 tasks. Within each, we draw $5$ architectures \emph{for each encoding} and show the mean $\pm$ 1 standard deviation (black and \textcolor{gray}{gray}, respectively) of the top-100 encodings vs those of \emph{all 1,000 sampled architectures} (\textcolor{red}{red}). Architectures sampled from the same encoding usually perform similarly and encodings that on average perform better also have smaller variability. 
    }    
    \label{fig:nasbench101_dist}
    \vspace{-4mm}
\end{figure*}

Knowing that most architectures perform similarly and that the true global optimum is very unlikely to be found, one can ask whether the current paradigm of searching for a specific architecture is the way forward. 
Differently from prior work, we argue that tackling \gls{NAS} with an approximate algorithm that learns a distribution rather than a specific architecture allows resources to be used more efficiently. An intuitive understanding can be had from Figure \ref{fig:example_archs}. 
Indeed, the accuracy distribution for each encoding is shown to have small standard deviation in Figure~\ref{fig:nasbench101_dist}, which motivates the optimization over encodings rather than architectures.
The key issue is that comparing every single architecture is intractable and unnecessary: intractable due to the sheer size of the considered search spaces, and unnecessary because we experimentally know that small differences in the architecture have little to no effect on the final result \cite{xie2019exploring,NEURIPS2020_NAGO,Yang2020NASEFH}.

Rather than searching for a specific architecture, we propose to find an approximate solution by instead searching for the optimal operation distribution, defined as the relative ratio between operations (e.g. \texttt{conv3$\times$3}, \texttt{conv5$\times$5},  \texttt{maxpool}, ...) in the whole architecture.
Once the operation distribution is defined, we can sample from it, seamlessly generating architectures of variable length. As we experimentally show, this simple re-framing of the \gls{NAS} problem enables us to significantly improve the sample efficiency of all exploitative methods. Indeed, \our is orthogonal to existing \gls{NAS} solutions and we experimentally show how it improves over Bayesian optimisation, local search and random search.
Not only is this approach more sample-efficient, but it has less potential for overfitting, as we show by successfully transferring to new datasets.

To summarise, we propose searching for an approximate rather than exact solution to the \gls{NAS} problem. This is easily applicable to most existing approaches and enables a substantial speed-up (from sample-efficiency) without a sacrifice in performance. 
We empirically show this both on the \gls{NAS} benchmarks \cite{dong2019NASbench201,siems2020bench} and with open-domain experiments on \gls{CIFAR-10} and \gls{CIFAR-100}.

\section{Related Work}

\textbf{Architecture encodings}
Several encoding schemes have been proposed to embed the directed acyclic graph of an cell-based architecture into a real-valued tensor to facilitate optimisation \cite{white2020study}. These encodings either represent the architecture via a flattened adjacency matrix and a list of operations \cite{ying2019bench} or characterise the architecture as a set of paths from input to output \cite{white2019bananas}. While multiple architectures may be projected to the same encoding and thus a particular encoding can also define a distribution of architectures, such performance standard deviation is rather small \cite{white2020study} and the search dimension remains high. On the other hand, some prior works use graph neural networks \cite{shi2020multiobjective}, graph kernels \cite{wan2021interpretable} or variational autoencoders \cite{luo2018NAO} to implicitly encode an architecture. These methods all require surrogate training and focus on comparing individual architectures rather than distributions of architectures.

\textbf{Stochastic methods} Recently, a number of methods based on probabilistic formulations have been proposed to overcome the memory requirements of differentiable approaches \cite{casale2019probabilistic,yan2020fp, MANAS2019}. These works learn a distribution over architectures and, at a first glance, might appear similar to what is being proposed in this work. The differences are significant and two-fold: (1) first and foremost the output of these methods is a  deterministic architecture and the distribution is learned for each possible placement of an operation---which does not reduce the search space size; and (2), their focus is strictly applied to the one-shot methods only. In other words, these works \cite{casale2019probabilistic,yan2020fp} propose a different search strategy rather than a different encoding, and as such are orthogonal to \our and could be combined with it. Another related class of works propose the concept of stochastic architecture generators \cite{xie2019exploring, NEURIPS2020_NAGO}. Specifically, they employ random graph generators to define the wiring pattern within an architecture and thus recast the problem of searching for an optimal architecture to finding the optimal generator hyper-parameters \cite{NEURIPS2020_NAGO}. This significantly reduces the dimension of the search space and improves the search efficiency. However, these prior works focus on the graph architecture topology, while keeping the operation type fixed. On the other hand, the focus of our work is to show that, regardless of any specific cell types, the operation distribution is an efficient and informative representation.

\section{Our Method: ANASOD}

\subsection{Overview}
\label{subsec:overview}

Searching for the optimal configuration of an architecture cell can be both extremely difficult (due to the formulation of \gls{NAS} as a combinatorial optimisation problem) and ineffective (due to the common practice of using different pipelines during search and evaluation and the imperfect correlation of performance between the two).
Instead, we propose to learn the optimal \emph{distribution of operations} within a cell. 
Formally, given an architecture cell $C$ with $N$ operation blocks $\{n_1,...,n_N\}$ where each operation is chosen from a pool of $\{o_1, ..., o_k\}$ candidate operations (e.g. in the \gls{NAS-Bench-201} cell, we have $N = 6$ and $k = 5$), its normalised distributional encoding (which we term the \textbf{\our} - \underline{A}pproximate \underline{\gls{NAS}} via \underline{O}peration \underline{D}istribution - encoding)  $\Tilde{\mathbf{p}}$ is a $k$-dimensional vector defined over the $k$-dimensional probability simplex $\{\Tilde{\mathbf{p}} \in \mathbb{R}^k, \sum_{i=1}^{k} \Tilde{p}_i = 1, \Tilde{p}_i \geq 0\; \forall i=\{1,\dots,k\}\}$ where $\Tilde{p}_i = \frac{n(o_i)}{N}$ and $n(o_i)$ is the number of blocks in the cell  $o_i$. The unnormalised encoding is $\mathbf{p} = N\Tilde{\mathbf{p}}$.

Because the encoding now only incorporates information about the operations in the cell, a single encoding maps to a \emph{distribution} of architectures sharing the same distribution of operations but possibly with different wiring. 
A key feature of \our is the easy reversibility in \emph{decoding} $\Tilde{\mathbf{p}}$ back to a distribution of architectures defined by it $p(\alpha|\Tilde{\mathbf{p}})$: Here we outline two possible ways to achieve so which are asymptotically equivalent when either the size of the cell becomes large, or when are sampling a large number of architectures. First, we may directly interpret the $i$-th element of the unnormalised encoding vector $\mathbf{p} = N\Tilde{\mathbf{p}}$ as the number of operation $i$ in the cell $C$; with the operations fixed, we may then randomly shuffle their ordering and the wiring amongst them to obtain a set of architectures $\{\alpha_1,\alpha_2,\dots\}$ 
This, however, requires $\sum_{i=1}^{k} p_i = N$ and $\mathbf{p} \in \mathbb{N}^k$.

Correspondingly, $\Tilde{\mathbf{p}}$ is only defined on a regular grid on the $k$-simplex, which could be problematic if we would like to apply standard continuous optimisation on the encoding space. To overcome this, we utilise the following rounding rule proposed in \cite{bomze2014rounding} to ``snap'' an arbitrary point on the simplex to a valid, all-integer configuration: denoting the standard $k$-simplex on which the \emph{unnormalised} encoding lies as $\Delta^k$ and an arbitrary point as $\mathbf{m}$. 
Define the fractional part of $\mathbf{m}$ as $\mathbf{s}(\mathbf{m}) = \big[m_i - \left \lfloor {m_i}\right \rfloor \big]_{i \in [1,k]}$, we have $g(\mathbf{m}) := \sum_{i=1}^k s(m_i) = N - \sum_{i=1}^k \left \lfloor {m_i}\right \rfloor \in [0, k-1]$ as a non-negative integer. We then round $g(\mathbf{m})$ largest elements of $s(m_i)$ to $1$ and the rest to $0$ to obtain a rounded integer vector $\mathbf{m}_r$:
\begin{equation}
\resizebox{0.45\textwidth}{!}{$
 {\mathbf {m_r}}:= \left[ \lfloor m_1\rfloor +1, \dots , \lfloor m_{g({\mathbf {m}})}\rfloor +1, \lfloor m_{g({\mathbf {m}})+1}\rfloor , \dots , \lfloor m_k\rfloor \right] ^\top$}
 \end{equation}
which, as \cite{bomze2014rounding} prove, is the closest valid configuration on $\Delta^k$ to $\mathbf{m}$ with respect to all $\ell^p$-norms for $p \geq 1$.

The second decoding method uses a probabilistic interpretation, and does not constrain $\Tilde{\mathbf{p}}$ to a finite set of points. At each operation block $n_j\; \forall j \in [1, N]$, we draw a sample from the categorical distribution parameterised by $\Tilde{\mathbf{p}}$:
\begin{equation}
    n_j \sim \mathrm{Cat}(\Tilde{p}_1, ... ,\Tilde{p}_k)\; \forall j \in [1, N]
\end{equation}
While the cells sampled via this decoding method only have operation distribution to be $\mathbf{\Tilde{p}}$ on expectation, in this case $\mathbf{\Tilde{p}}$ is not constrained to the grid like before, enabling gradient-based continuous optimisation to be used effectively and as we show in Sec. \ref{subsec:applyinganasod}, is key for applying \our in the differentiable \gls{NAS} setting.

\begin{table}[!t]
    \centering
    \caption{Overall \gls{SD}, median \gls{SD} of \emph{different} architectures sampled from the \emph{same encoding} and the median \gls{SD} of the \emph{same} architecture trained with \emph{different seeds}. All \gls{SD} are w.r.t validation error in percentage.}
    \label{tab:std}
\begin{footnotesize}
\begin{tabular}{@{}lcccc@{}}
\toprule
Benchmark & \multicolumn{3}{l}{\gls{NAS-Bench-201}} & \textsc{nb301} \\
Task & C10 & C100 & ImageNet16 & C10 \\
\midrule
Overall & $9.5$ & $14$ & $10$ & $0.77$\\
\midrule
Median (same encoding) \\
 - All encodings &$1.2$ & $2.4$ & $2.7$ & $0.25$\\
 - Top $50\%$-performing & $0.84$ & $1.4$ & $1.9$ & $0.22$ \\ 
 \midrule
Median (different seeds) & $0.19$ & $0.35$ & $0.36$ & $0.17$\\
\bottomrule
\end{tabular}
\end{footnotesize}
\end{table}

For the proposed encoding to provide a reasonable approximation of the \gls{NAS} solution, it is imperative for the architectures sampled from the same encoding to perform similarly, which we empirically show to be the case in the popular cell-based search spaces represented by the \gls{NAS-Bench-201} \cite{dong2019NASbench201} and \gls{NAS-Bench-301} \cite{siems2020bench} datasets. With reference to Figure \ref{fig:nasbench101_dist} and Table \ref{tab:std}, in each of the $4$ tasks considered, we draw $200$ random encoding $\Tilde{\mathbf{p}}$ and within each encoding, we draw $5$ unique architectures sampled from that encoding and query the validation performance of all $1,000$ architectures. To investigate the extent to which \our encoding determines the validation performance, we compute the overall \gls{SD} across all sampled architectures which acts as an unbiased estimate of the performance variability over the entire search space, We also consider the median \gls{SD} amongst architectures sharing the same encoding, with particular attention to the better-performing encodings in which \gls{NAS} is particularly interested. Additionally, we also report the median \gls{SD} of the same architectures trained with different seeds, representing the amount of irreducible aleatory uncertainty due to evaluation noise. It is evident that fixing the \our encoding massively reduces the validation error variability and hence is highly predictive of the performance, and comparatively, better encodings have lower variability.

Having established the \emph{validity} of \our, we now argue there are 3 key \emph{advantages} of \our. First, it requires no learning and massively compresses the search space, yet remains predictive enough about the performance of the architectures. While as discussed, the original architecture space explodes exponentially as $k^N$, the number of unique \our encoding is simply $\binom{N+k-1}{k-1}$.  
For instance, in the \gls{DARTS}-like cell with $k=8$ and $N=14$, the number of architectures is $8^{14}\approx 4.3 \times 10^{12}$, but the number of \our encodings is only $\binom{21}{7} \approx 1.2 \times 10^{5}$. This enables us to reduce the difficult combinatorial \gls{NAS} optimisation into a much more tractable, easier-to-solve problem and thus to achieve superior sample efficiency -- this is a result of \our being an approximate encoding; exact encoding such as the adjacency or path encoding, while representing the architecture in a way that is amenable for predictor learning, usually do not reduce the optimisation difficulty \textit{per se}. Second, as mentioned, we may generatively sample architectures from the encoding easily. This allows us to directly search for an optimal encoding $\Tilde{\mathbf{p}}^*$ first and to map back to the architectures thereafter for effective (approximate) global optimisation over the entire search space; this is in contrast with many previous approaches which, although using advanced encoding schemes, still rely on random sampling and/or mutation algorithm in the \emph{architecture} space for architecture optimisation \cite{wan2021interpretable,shi2020multiobjective,white2019bananas}. 
Lastly, unlike most existing encodings, \our is defined on a well-studied simplex space with well-defined distance metrics and is of a modest dimensionality; all of which are key for effective deployment of model-based methods such as \gls{GP}-based Bayesian optimisers, which are often difficult to apply \gls{NAS} due to the high-dimensional search spaces and/or the lack of natural distance metrics.

\subsection{Applying ANASOD}
\label{subsec:applyinganasod}

In this section, we give some examples that orthogonally combine various methods with \our to achieve improved efficiency, performance, or both. It is worth emphasising that the methods proposed here are exemplary, and in no way exhaust the applicability of the proposed encoding.

\paragraph{\gls{RS}} While simplistic in implementation, \gls{RS} is shown to be a remarkably strong baseline in \gls{NAS} where even the state-of-the-art optimisers often fail to significantly outperform \gls{RS} \cite{ying2019bench, dong2019NASbench201,Yang2020NASEFH}. Implementing random sampling in the proposed encoding space is trivial: to sample uniformly from the entire search space, we simply draw encoding from the uniform Dirichlet distribution $\Tilde{\mathbf{p}}_i \sim \mathrm{Dir}(1, ..., 1)$. 
More importantly, by virtue of the unique property that our encoding lies on a probability simplex on which well-principled distributions can be easily defined, we also propose a simple modification that mitigates the key shortfall of \gls{RS}, the fact that it is purely exploratory and does not exploit the information gathered as the optimisation progresses: at iteration $t$, instead of randomly drawing the next encoding $\Tilde{\mathbf{p}}_{t+1}$ from the uniform distribution, we instead draw randomly the next sample from the following distribution whose mode is $\Tilde{\mathbf{p}}^*$, the best-performing encoding seen so far:
\begin{equation}
    \Tilde{\mathbf{p}}_{t+1} \sim \mathrm{Dir}(\alpha_1, ..., \alpha_k) \text{ where } \alpha_i = k\beta_t\Tilde{p}^*_i  + 1 \text{ } \forall i,
    \label{eq:biasedrs}
\end{equation}
where $\beta_t$ is a parameter similar to the temperature parameter in simulated annealing that monotonously increases from $0$ (the uniform Dirichlet) to some positive $\beta_T$ at termination of the search that controls the scale of the Dirichlet distribution. Essentially, in this modified \gls{RS} (which we term \emph{biased \gls{RS}}), we add exploitation by progressively biasing the generating distribution of the encoding to the locality around the best encoding seen so far with the hope of finding even better ones, with $\beta_t$ controlling the amount of exploitation-exploration trade-off. This modification is simple, poses almost no additional computing cost (apart from the need to track $\Tilde{\mathbf{p}}^*$ as the search progresses), and is orthogonal to other modifications of \gls{RS} (e.g. \gls{RS} with successive halving \cite{li2016hyperband} and \gls{RS} with parameter sharing \cite{li2020random}), but as we will show in Sec. 4 it significantly improves on the vanilla \gls{RS} on a wide range of search spaces to a level that often matches or even exceeds more sophisticated methods, and thus could serve as a strong baseline for the \gls{NAS} community. 

\paragraph{Differentiable NAS} Another popular paradigm in \gls{NAS} is that of \emph{\gls{DNAS}} pioneered by \cite{Liu2019_DARTS}, which, generally speaking, utilises \emph{continuous relaxation} of the discrete architecture space (followed by \emph{discretisation} at the end of the pipeline) with weight-sharing supernets so that standard first-order optimisers commonly used in deep learning might be applied in this context. Differentiable \gls{NAS} finds reasonable solutions, and often offers a massive speedup compared to earlier query-based methods, but it is also subjected various criticisms, such as the so-called \emph{rank disorder} between the found cells during search and evaluation and the unstable training with a propensity to collapse into cells with parameterless operations that corresponds to poorly-performing architectures \cite{Yang2020NASEFH, zela2019understanding}
In this section, we describe a prototypical application of \our in the context of \gls{DNAS}, and shows that it performs competitively and leads to further saving in computation over the already efficient \gls{DNAS} pipeline while remaining much simpler than many of the existing methods.

To apply \our in the differentiable setting, we first notice that the stochastic interpretation of the \our encoding (described in Section \ref{subsec:overview}) as the parameters governing the categorical distribution at \emph{each} operation block $j \in [1, N]$ of a cell. As described, this relaxes the constraint that \our encoding is only defined on a regular grid (the encoding only needs to obey $\sum_i \Tilde{p}_i = 1$). With this, we might compute the derivative of the supernet validation loss $\mathcal{L}_{\mathrm{val}}(\Tilde{\mathbf{p}}, w)$ w.r.t the \emph{encoding},  $\nabla_{\Tilde{\mathbf{p}}}\Big(\mathcal{L}_{\mathrm{val}}(\Tilde{\mathbf{p}}, w)\Big)$ where $w$ is the supernet parameter weights. We formulate the procedure as a bi-level optimisation problem like \cite{Liu2019_DARTS}, and include an algorithmic description in Algorithm \ref{alg:diffNAS}.

\begin{algorithm}[!ht]
\begin{footnotesize}
	    \caption{\our-\gls{DNAS}. Key differences from existing \gls{DNAS} algorithms marked \textcolor{blue}{blue}.}
	    \label{alg:diffNAS}
	\begin{algorithmic}[1]
	\STATE Create a mixed operation $\Bar{o}^j \sim \mathrm{Cat}(\Tilde{\mathbf{p}})$ for each operation block. \textcolor{blue}{Note that the parameters of the categorical distribution are shared.}
	\WHILE{not converged}
	\STATE Update the \textcolor{blue}{encoding $\Tilde{\mathbf{p}}$} by descending $\nabla_{\Tilde{\mathbf{p}}}\Big(\mathcal{L}_{\mathrm{val}}(\Tilde{\mathbf{p}}, w)\Big)$, and keep $w$ constant 
	\STATE \textcolor{blue}{Enforce the simplex constraint $\Tilde{\mathbf{p}} \leftarrow \frac{\Tilde{\mathbf{p}} }{\sum_i^k \Tilde{p_i} }$ (i.e. mirror descent)}
	\STATE Update the supernet weights $w$ by descending $\nabla_w \mathcal{L}_{\mathrm{train}}(\Tilde{\mathbf{p}}, w)$, and keep $\Tilde{\mathbf{p}}$ constant.
	\ENDWHILE
	\STATE \textcolor{blue}{Sample cells from the optimised encoding $\alpha \sim p(\alpha|\Tilde{\mathbf{p}}^*)$ and stack them into a final neural architecture.}
	\end{algorithmic}
\end{footnotesize}
\end{algorithm}

While we inherit some key elements from the existing \gls{DNAS} algorithms such as the idea of bi-level optimisation and most hyperparameter choices (e.g. the learning rate of the architecture optimiser and the network architecture), there are two key differences: first, inspired by our fundamental observation that often determining the operation mix is already sufficiently predictive of the architecture performance, we use the $k$-dimensional \our encoding of the cell as the parameters of the single categorical distribution that governs the all edges in the cell; on the other hand, existing methods uses a $k$-dimensional vector at each edge, leading to a matrix parameterisation of the cell of dimensionality $N \times k$; as we will show later, we argue that our simpler representation implicitly regularises the architecture learning and avoids the catastrophic collapse of many \gls{DNAS} methods due to over-fitting; at the same time, the reduced dimensionality allows fast convergence as we will demonstrate. Second, instead of taking $\arg \max$ of the categorical distribution parameters on each edge at the end of the search, which is hypothesised as a key reason leading to rank disorder (as the discretised architecture might differ a lot from the continuously relaxed architecture in the parameter space \cite{Yang2020NASEFH, yu2019evaluating, zela2019understanding}), we simply sample architecture cells from the \our-encoding at the end to build a final network.

\paragraph{\gls{LS}} At the other end of the exploration-exploitation continuum, \gls{LS} is purely exploitative, and has recently been shown to be an extremely powerful optimiser, achieving state-of-the-art performance in small search spaces but fails to perform on larger ones \cite{white2020local}. We show that combining \our with \gls{LS} universally further improves performance on \gls{LS} but particularly in the latter case, addressing the previous failure mode of \gls{LS}. The readers are referred to App. \ref{app:local_search} for a detailed discussion.

\paragraph{\gls{SMBO}} \gls{SMBO} algorithms, in particular \gls{BO}, have seen great success in hyperparameter optimisation due to their exploitation-exploration balancing, sample efficiency, and principled treatment of and robustness against noises that inevitably arise in real life. Nonetheless, their application in \gls{NAS} has been somewhat constrained, partly due to the discrete search space, extreme dimensionality of the existing encoding and the common lack of well-defined distances in such space (the presence of which is crucial for kernel-based methods, such as \gls{GP}). 
As discussed, \our addresses all the three aforementioned issues and enables \gls{GP}-based \gls{BO} to be effectively used. \our-\gls{BO}  is described in Algorithm \ref{alg:bo}. Nonetheless, we emphasise that \our-\gls{BO} is not restricted to the \gls{GP} surrogate; as we show in Sec. 4, it may also be used with \gls{NEP} proposed in \cite{white2019bananas} to deliver further performance improvements.

For easy parallelism which is key in \gls{NAS}, we opt for a sampling-based optimisation of the acquisition function which allows trivial modifications to be able to recommend batches of points simultaneously, although we are nonetheless compatible with, for example, the common gradient-based optimiser. 
On this, in addition to the usual \gls{BO} components, inspired by the recent \emph{trust region}-based innovations in \gls{BO} literature \cite{eriksson2019scalable, wan2021think} we also adopt dynamic adjustment of the \emph{encoding generating distribution} (Lines 2 and 9 in Algorithm \ref{alg:bo}). 
Specifically, similar to \our-\gls{RS}, we use the knowledge gained previously (i.e. the location of the best encoding so far $\Tilde{\mathbf{p}}^*_t$ as a prior to generate samples for the next iteration by biasing the Dirichlet distribution. 
However, instead of using a simple annealing schedule for the $\beta$ parameter, here we adopt a probabilistic trust-region setting, where we halve $\beta$ (i.e. reducing the concentration of the Dirichlet density around $\Tilde{\mathbf{p}}^*$ to allow candidate encodings to be drawn from a larger region) upon successive successes in reducing the best function value seen so far $y^*_t$ up to 0.\footnote{In \cite{eriksson2019scalable}, the trust regions are set as ``hard'' hyper-rectangular bounding boxes around $\Tilde{\mathbf{p}}^*_t$. Here, we place a ``soft'', high-probability trust region by adjusting $\beta$ to constrain most of the probability mass within the locality.} Conversely, we double $\beta$, which is equivalent to increasing the concentration of the Dirichlet p.d.f around the mode and thus to shrink the trust region size upon successive failures, up to a certain $\beta_{\max}$.

\begin{algorithm}[!t]
\begin{footnotesize}
	    \caption{\our-\gls{BO}. Key differences from conventional \gls{BO} are marked \textcolor{blue}{blue}.}
	    \label{alg:bo}
	\begin{algorithmic}[1]
		\STATE {\bfseries Input:} Objective function (default: validation error) $y$, number of initialising random samples $n_{\mathrm{init}}$
		\STATE Initialise the \emph{encoding generating distribution} to the uniform Dirichlet distribution $p(\Tilde{\mathbf{p}}) = \mathrm{Dir}(1, ..., 1)$
		\STATE Sample $n_{\mathrm{init}}$ random encodings $\Tilde{\mathbf{p}}_{[1:n_{\mathrm{init}}]} \sim p(\Tilde{\mathbf{p}})$ and evaluate to obtain $y(\Tilde{\mathbf{p}})$ to initialise the surrogate \gls{GP}.
		\FOR{i=$n_{\mathrm{init}}$,,,.T}
		\STATE Sample a pool of $B$ candidate encodings from $p(\Tilde{\mathbf{p}})$
		\STATE Select the next query point(s) by identifying the encoding that maximises the acquisition function $\Tilde{\mathbf{p}}_{i} = \arg \max \big( \mathrm{acq}(\Tilde{\mathbf{p}}) \big)$.
		\STATE \textcolor{blue}{Evaluate a single architecture $\alpha_i$ from the encoding $\Tilde{\mathbf{p}}_{i}$ to approximate the performance of all architectures parameterised by $\Tilde{\mathbf{p}}_{i}$}.
		\STATE Augment the surrogate \gls{GP} with new encoding-observation pair(s) $\mathcal{D}_i \leftarrow \mathcal{D}_{i-1} \cup \{\Tilde{\mathbf{p}}_{i}, y(\Tilde{\mathbf{p}}_i)\}$ and optimise the \gls{GP} hyperparameters via the marginal log-likelihood maximisation.
		\STATE Update the \textcolor{blue}{encoding generating distribution $p(\Tilde{\mathbf{p}})$.}
		\ENDFOR
	\end{algorithmic}
\end{footnotesize}
\end{algorithm}

\begin{figure*}[t]
    \vspace{-4mm}
    \centering
     \begin{subfigure}{0.23\linewidth}
    \includegraphics[trim=0cm 0cm 0cm  0cm, clip, width=1.0\linewidth]{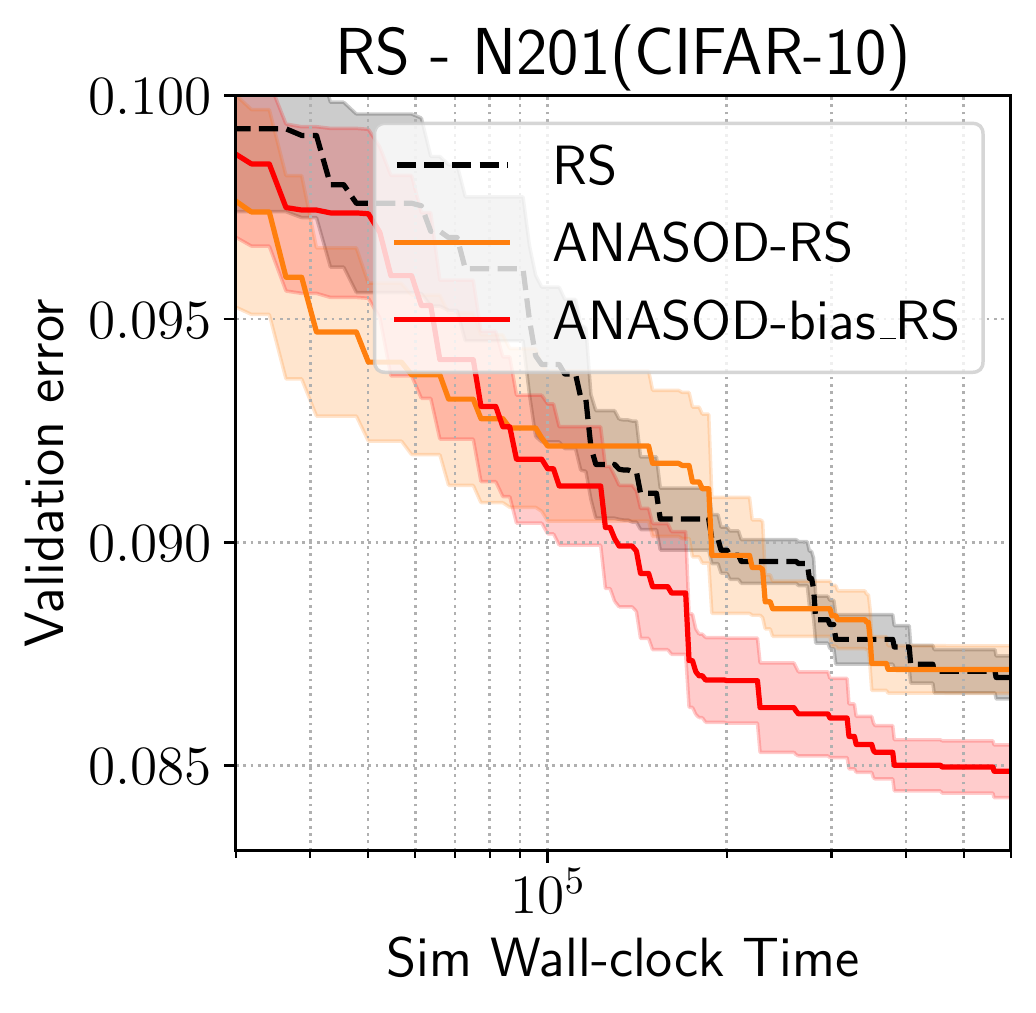}
    \end{subfigure}
    \begin{subfigure}{0.23\linewidth}
        \includegraphics[trim=0cm 0cm 0cm  0cm, clip, width=1.0\linewidth]{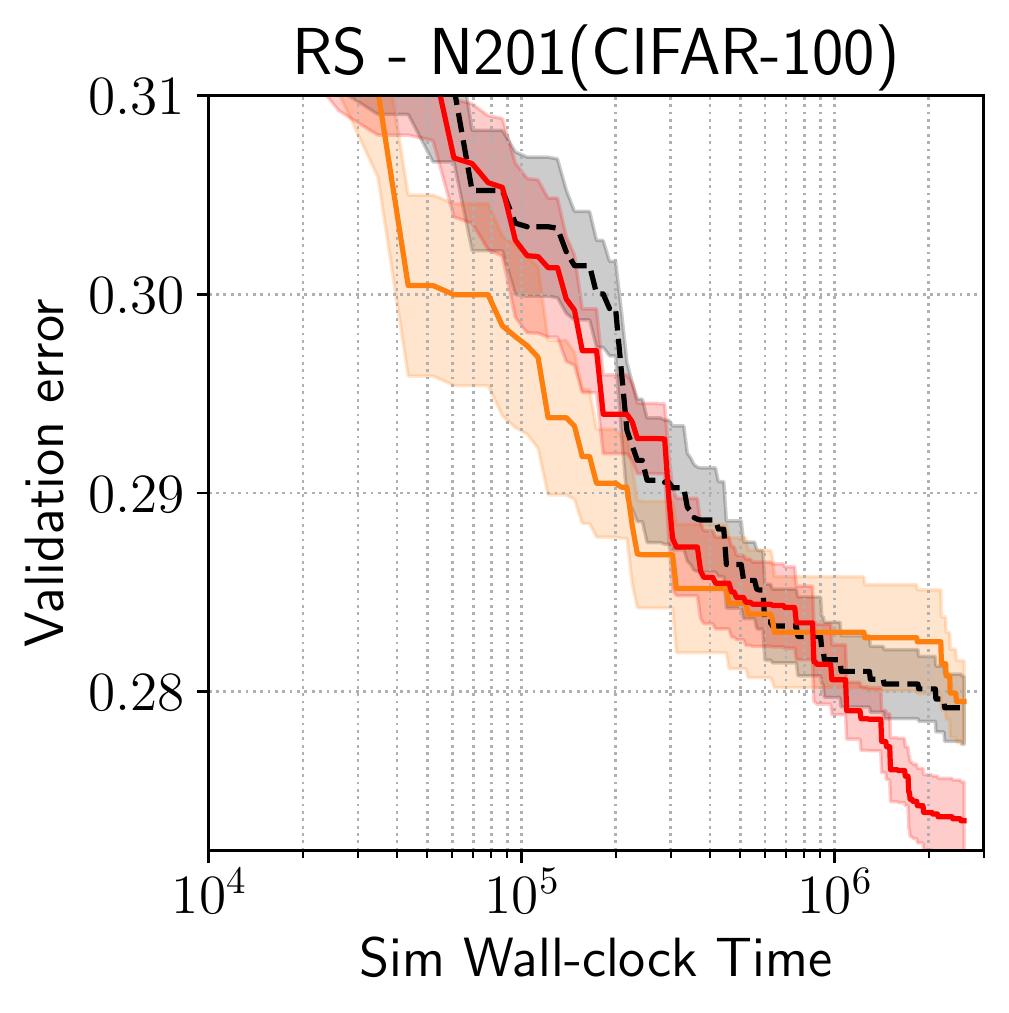}
    \end{subfigure}
         \begin{subfigure}{0.23\linewidth}
    \includegraphics[trim=0cm 0cm 0cm  0cm, clip, width=1.0\linewidth]{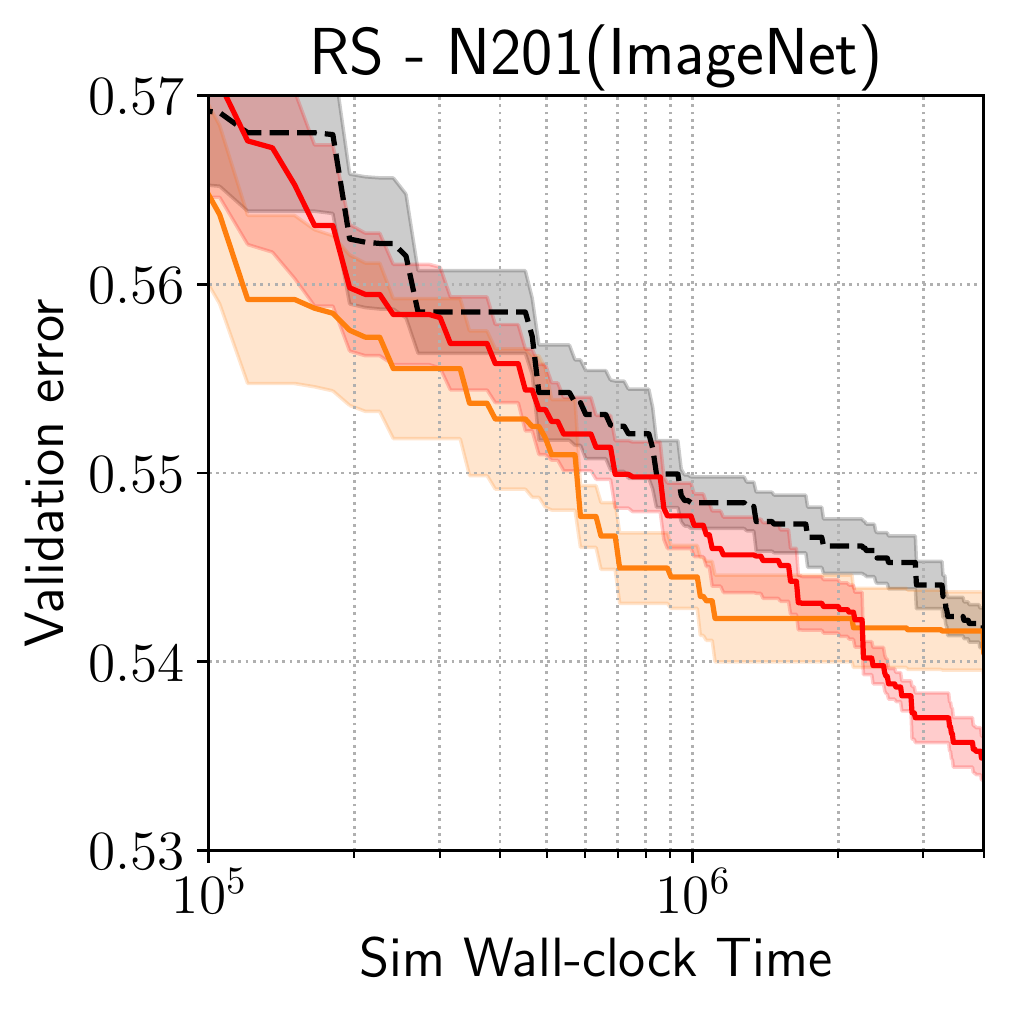}
    \end{subfigure}
             \begin{subfigure}{0.23\linewidth}
    \includegraphics[trim=0cm 0cm 0cm  0cm, clip, width=1.0\linewidth]{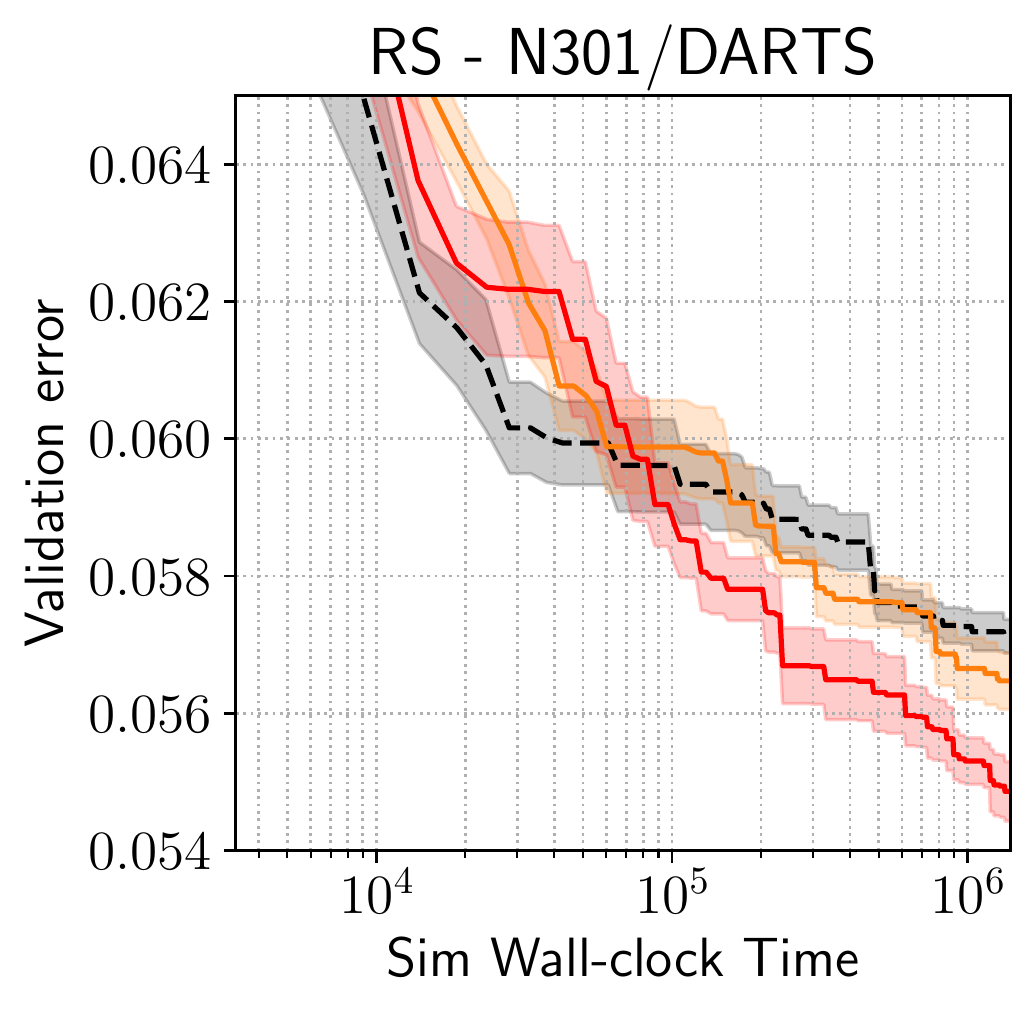}
    \end{subfigure}

             \begin{subfigure}{0.23\linewidth}
    \includegraphics[trim=0cm 0cm 0cm  0cm, clip, width=1.0\linewidth]{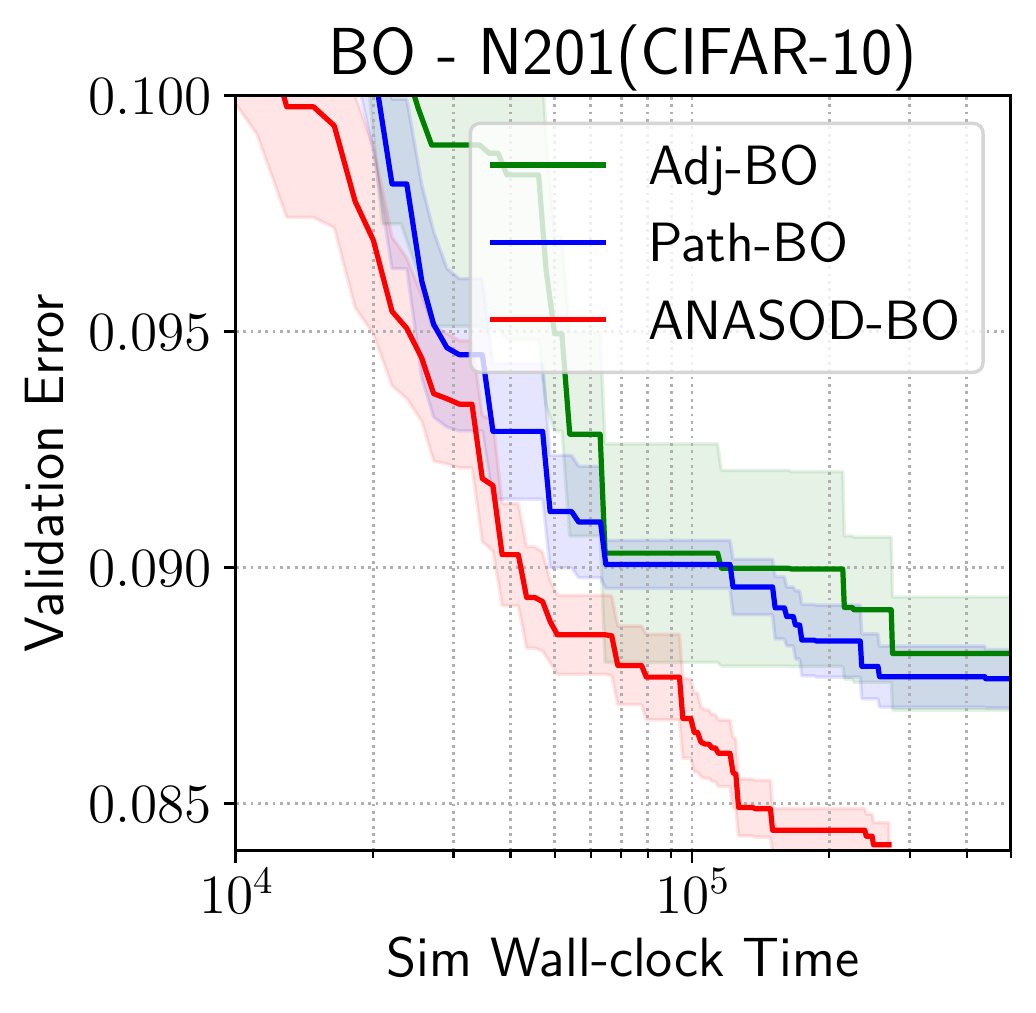}
    \end{subfigure}
    \begin{subfigure}{0.23\linewidth}
        \includegraphics[trim=0cm 0cm 0cm  0cm, clip, width=1.0\linewidth]{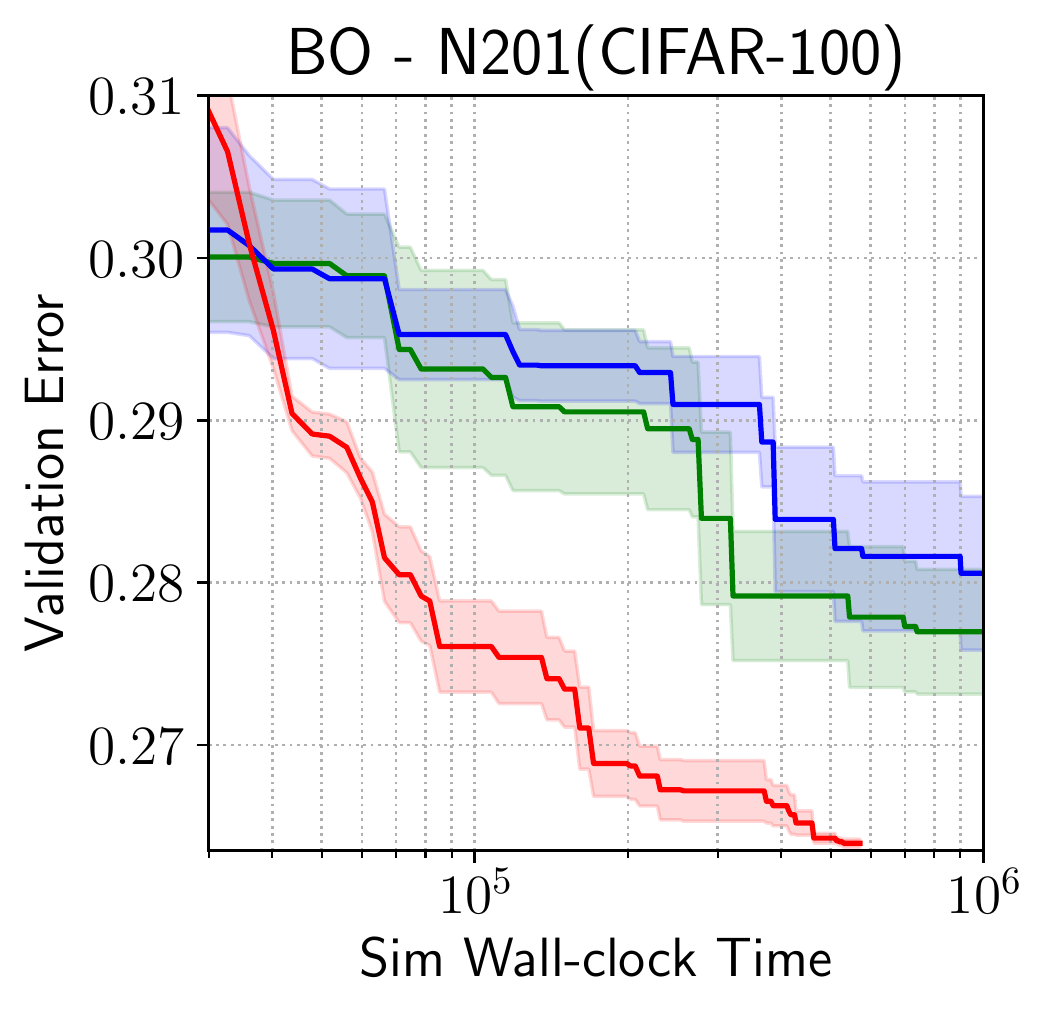}
    \end{subfigure}
         \begin{subfigure}{0.23\linewidth}
    \includegraphics[trim=0cm 0cm 0cm  0cm, clip, width=1.0\linewidth]{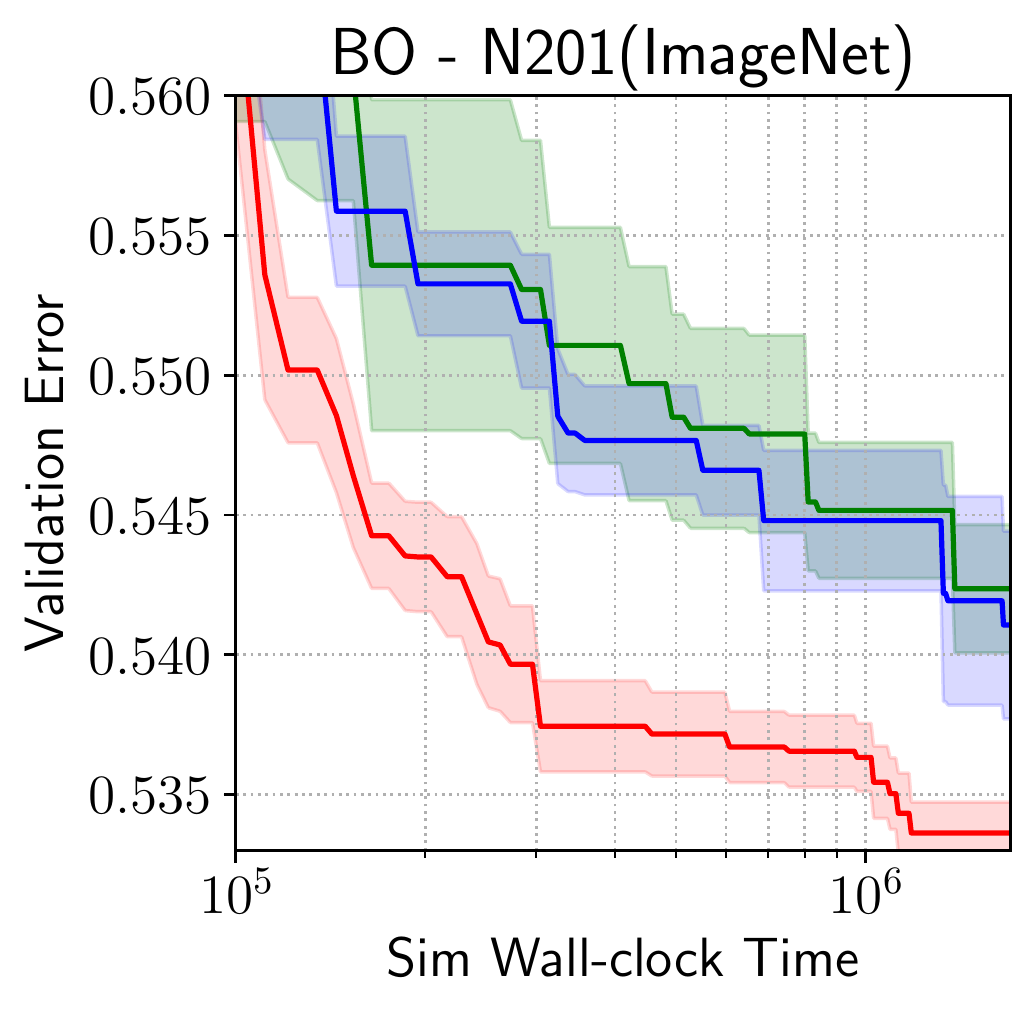}
    \end{subfigure}
    \begin{subfigure}{0.23\linewidth}
    \includegraphics[trim=0cm 0cm 0cm  0cm, clip, width=1.0\linewidth]{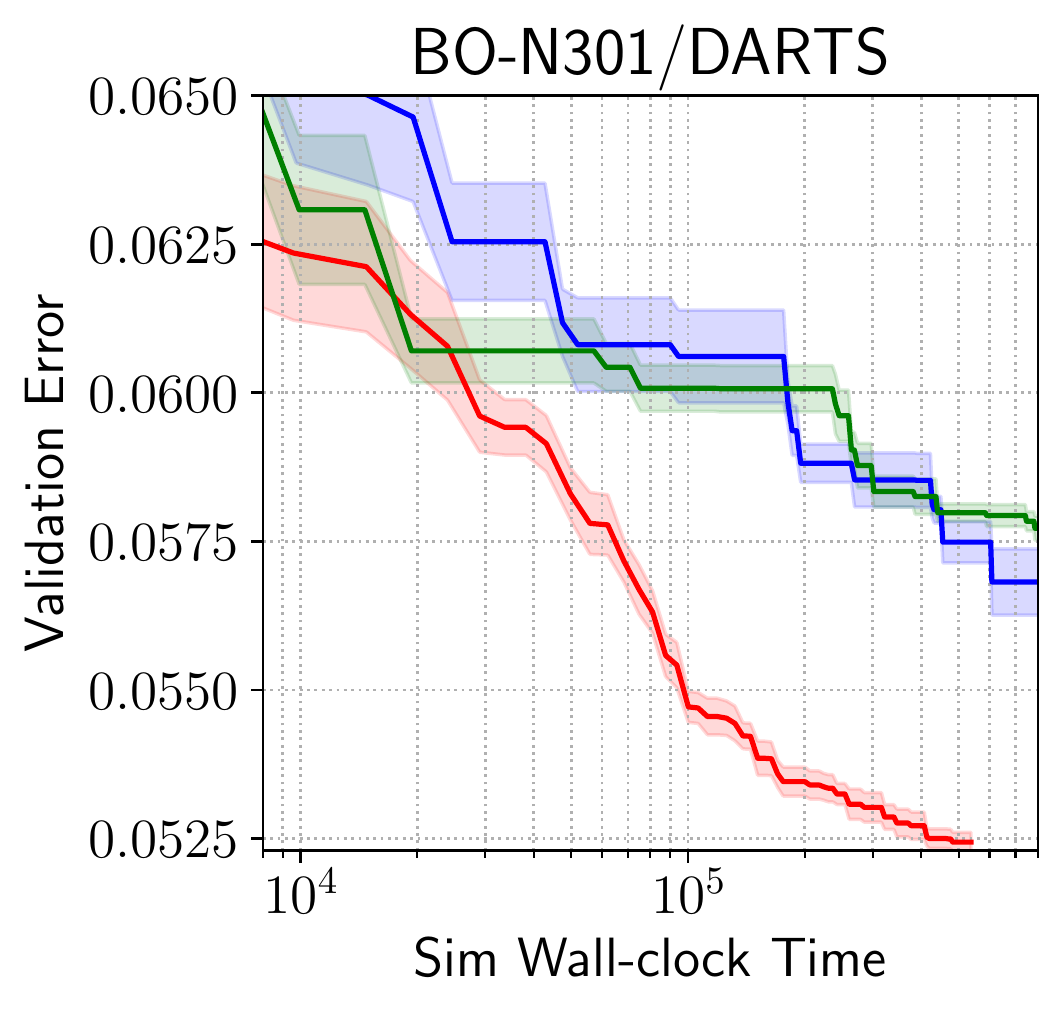}
    \end{subfigure}
    
        \begin{subfigure}{0.23\linewidth}
        \includegraphics[trim=0cm 0cm 0cm  0cm, clip, width=1.0\linewidth]{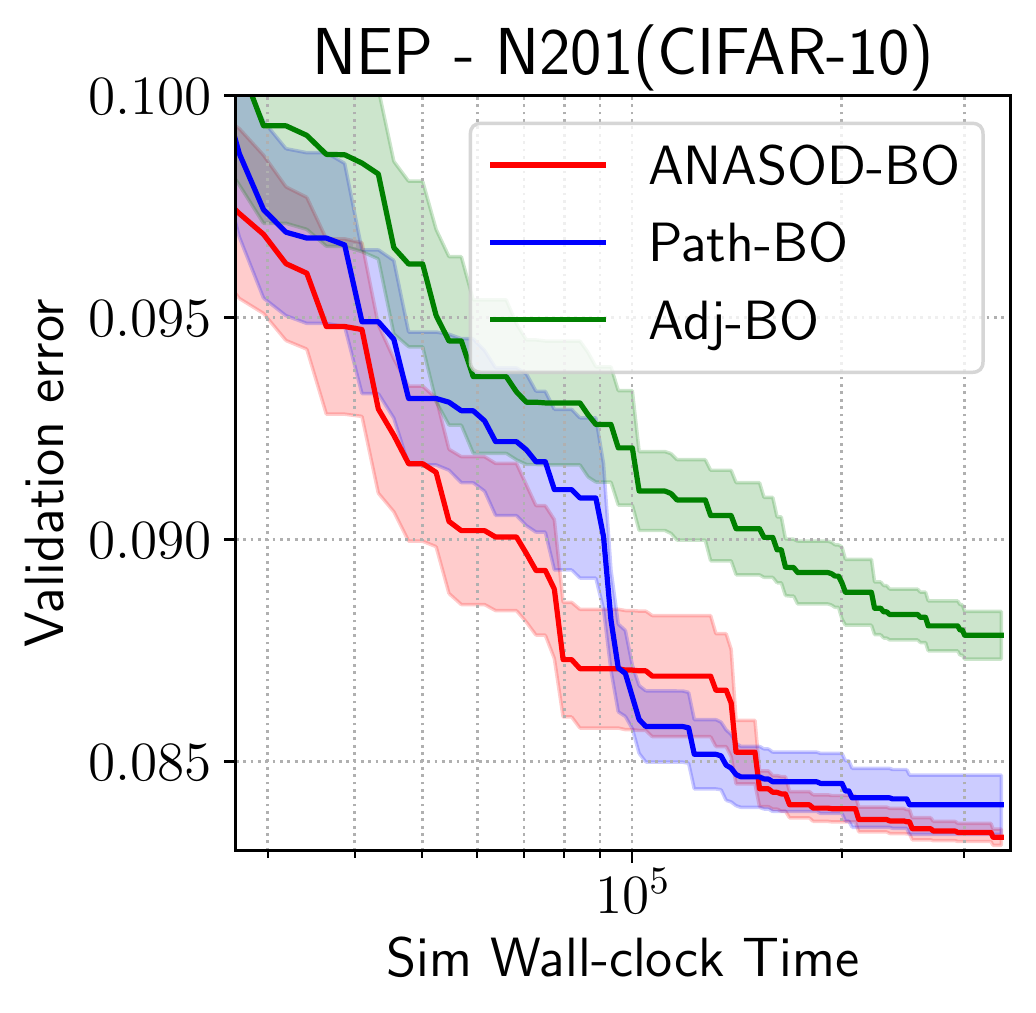}
    \end{subfigure}
    \begin{subfigure}{0.23\linewidth}
        \includegraphics[trim=0cm 0cm 0cm  0cm, clip, width=1.0\linewidth]{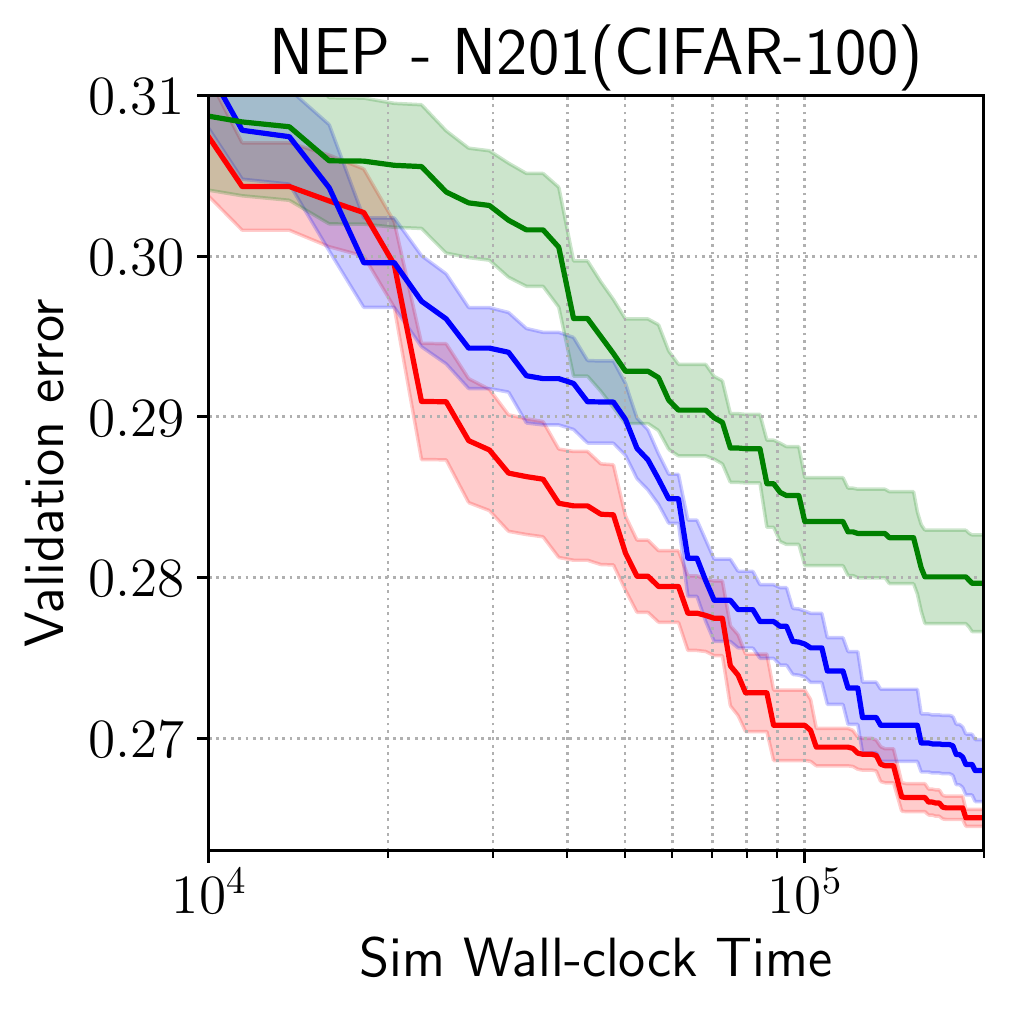}
    \end{subfigure}
        \begin{subfigure}{0.23\linewidth}
        \includegraphics[trim=0cm 0cm 0cm  0cm, clip, width=1.0\linewidth]{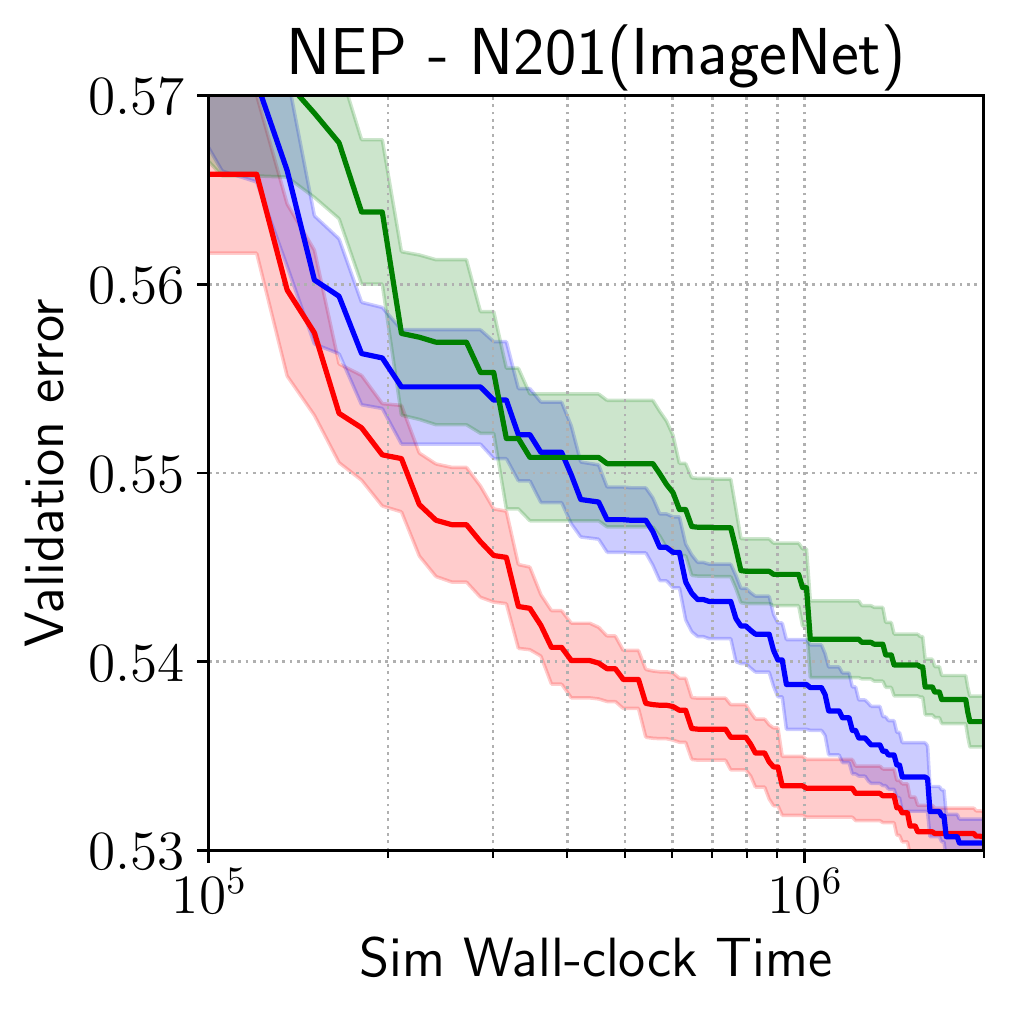}
    \end{subfigure}
        \begin{subfigure}{0.23\linewidth}
        \includegraphics[trim=0cm 0cm 0cm  0cm, clip, width=1.0\linewidth]{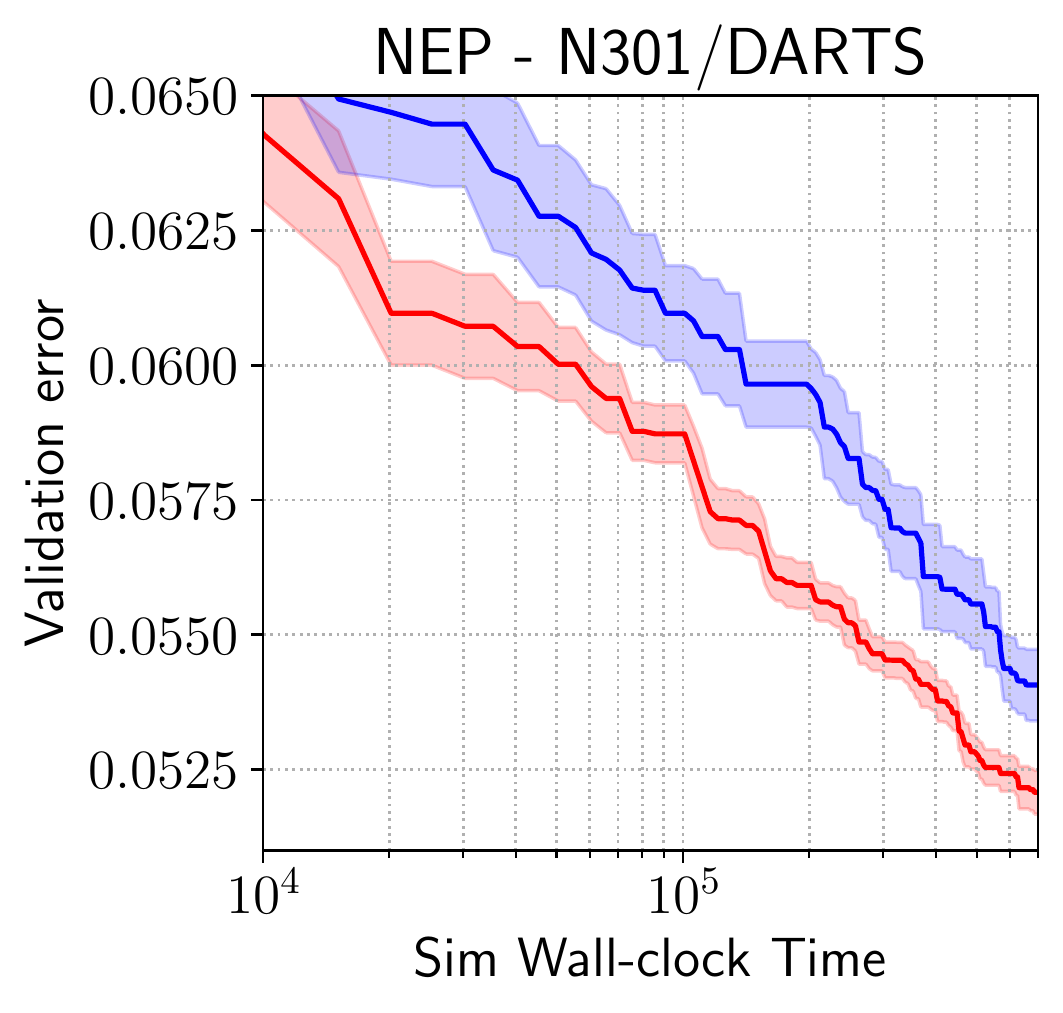}
    \end{subfigure}
        \vspace{-4mm}
    \caption{Performance of various methods on \gls{NAS-Bench-201} and \gls{NAS-Bench-301} with and without the \our encoding: (\textbf{Top row}) random search, \gls{RS}; 
    ({\bfseries Middle row}) \gls{GP}-\gls{BO};  ({\bfseries Bottom row}) \gls{NEP}-\gls{BO}. Note the x-axis which shows the (simulated) \textsc{gpu}-seconds is in log-scale. 
    For \gls{RS} %
    we set a budget of maximum 300 architecture queries; for \gls{BO}, we set a more stringent budget of 150 queries for \our-\gls{BO} but allow the baselines to run for longer (to observe the amount of speedup). Adj-\gls{BO} and Path-\gls{BO} are the variants of \gls{BO} that are otherwise identical to \our-\gls{BO} as outlined in Algorithm \ref{alg:bo}, but with the \our encoding replaced by the adjacency \cite{ZophLe17_NAS, ying2019bench} and path \cite{white2019bananas, wei2020npenas} encoding, respectively. Lines and shades denote mean $\pm$ 1 standard error, across 10 different random trials.
    }    
    \label{fig:nasbench}
    \vspace{-4mm}
\end{figure*}

Since we conduct \gls{BO} on the encoding space directly, at termination \our-\gls{BO} returns the optimised encoding $\Tilde{\mathbf{p}}^*$ instead of a single architecture $\alpha^*$. This may be of independent benefits: for example, the \our encoding, which has much fewer parameters than an exact architecture cell and its exact encoding thereof, should be much less prone to overfitting to a singular data-set which is cited as a problem many \gls{NAS} methods have \cite{lian2019towards}. Furthermore, since multiple architectures are sampled from a single encoding, we might also obtain an \emph{ensemble} of architectures which is recently shown to perform very competitively both in terms of accuracy and uncertainty quantification \cite{ardywibowo2020nads, zaidi2020neural}. We defer a thorough investigation to a future work.

\section{Experiments}

\begin{table}[!t]
    \centering
    \caption{Performance on NAS-Bench datasets. Unless otherwise specified, we report mean $\pm$ 1 standard error of the validation error across 10 random trials. For fair comparison, in \gls{RS} and \gls{LS} experiments, the numbers shown denote the best validation error seen after $\mathbf{300}$ architecture queries; in \gls{BO} experiments, we show the best validation error seen for each method after $\mathbf{150}$ architecture queries (which is the budget we set for \our-\gls{BO}).}
\begin{footnotesize}
\resizebox{\columnwidth}{!}{%
\begin{tabular}{@{}lcccc@{}}
\toprule
Benchmark & \multicolumn{3}{c}{\gls{NAS-Bench-201}} & \gls{NAS-Bench-301} \\
Dataset & \gls{CIFAR-10} & \gls{CIFAR-100} & ImageNet16 & \gls{CIFAR-10} \\
\midrule
\gls{RS} & $8.67_{\pm 0.03}$ & $27.91_{\pm 0.17}$ & $54.17_{\pm 0.10}$ & $5.71_{\pm 0.02}$\\
\textbf{\our-\gls{RS}} & $8.71_{\pm 0.05}$ & $27.95_{\pm 0.20}$ & $53.85_{\pm 0.12}$ & $5.65_{\pm 0.04}$\\
\textbf{\our-biased\gls{RS}} & $\mathbf{8.48}_{\pm 0.06}$ & $\mathbf{27.35}_{\pm 0.20}$ & $\mathbf{53.30}_{\pm 0.10}$ & $\mathbf{5.47}_{\pm 0.04}$\\
\midrule
RL \cite{ZophLe17_NAS} & $8.91_{\pm 0.05}$ & $28.15_{\pm 0.18}$ & $54.45_{\pm 0.12}$ & *\\
RE \cite{real2019regularized} & $8.86_{\pm 0.05}$ & $28.40_{\pm 0.14}$ & $54.28_{\pm 0.10}$ & $5.62_{\pm 0.03}$\\
SMAC \cite{hutter2011sequential} & $8.89_{\pm 0.05}$ & $27.80_{\pm 0.20}$ & $53.64_{\pm0.13}$ & $5.45_{\pm 0.03}$\\
TPE \cite{bergstra2011algorithms} & $8.57_{\pm 0.04}$ & $27.28_{\pm 0.14}$ & $53.54_{\pm0.14}$ & $5.51_{\pm 0.02}$\\
GCNBO \cite{shi2020multiobjective} & $8.84_{\pm 0.01}$ & $27.93_{\pm 0.03}$ & $53.46_{\pm 0.06}$ & $5.54_{\pm 0.04}$\\
BANANAS \cite{white2019bananas} & $8.51_{\pm 0.08}$ & $26.53_{\pm 0.02}$ &$53.41_{\pm 0.04}$ & $5.36_{\pm 0.05}$\\
NAS-BOWL \cite{wan2021interpretable} & $8.50_{\pm 0.09}$ & $26.51_{\pm 0.00}$ & $\mathbf{53.36}_{\pm 0.04}$ & $5.31_{\pm 0.06}$\\
\textbf{\our-\gls{BO}} & $\mathbf{8.41}_{\pm 0.05}$ & $\mathbf{26.41}_{\pm 0.02}$ & $\mathbf{53.36}_{\pm 0.10}$ & $\mathbf{5.24}_{\pm 0.02}$\\
\bottomrule
\multicolumn{5}{l}{$^{*}$: The original repo does not support the \gls{NAS-Bench-301} search space.}
    \end{tabular}
    }
\end{footnotesize}

    \label{tab:nasbench}
\end{table}

\subsection{Experiments on the NAS-Bench Datasets} 
\label{subsec:nasbench}
We first conduct optimisation experiments on the \gls{NAS-Bench-201} and \gls{NAS-Bench-301} datasets \cite{dong2019NASbench201,siems2020bench}. On \gls{NAS-Bench-201}, we consider all three tasks: \gls{CIFAR-10}, \gls{CIFAR-100} and ImageNet16. 
\gls{NAS-Bench-301} is a surrogate-based benchmark that is trained on \gls{CIFAR-10} only, but it contains as many distinct architectures as the \gls{DARTS} search space and is thus much more challenging. We do not conduct experiment on the earlier \gls{NAS-Bench-101} \cite{dong2019NASbench201}, as similarly to \gls{NAS-Bench-301} it is also only trained on \gls{CIFAR-10}, but its search space ($\sim 4 \times 10^5$ architectures) is much smaller.

\paragraph{Comparison with baseline strategies}

To enable direct and fair comparison of algorithms within each class, we first compare \gls{RS}, \gls{GP}-\gls{BO} and \gls{NEP}-\gls{BO} against their \our counterpart, as described in Section~\ref{subsec:applyinganasod}. 

In the top row of Figure~\ref{fig:nasbench}, we observe that \our-\gls{RS} performs on par with \gls{RS}---this is unsurprising, as both methods sample uniformly from the same, entire search space which is independent of the specific encoding chosen to represent it. 
This result also acts as a sanity check that there is no noticeable bias in sampling in the \our space as contrasted to the original architecture space. On the other hand, \our-\gls{RS} with bias, which randomly samples from a non-uniform Dirichlet distribution, can be seen to significantly outperform them both as the optimisation progresses, across all tasks, further highlighting its potential as a simple yet robust baseline in \gls{NAS} research.

The middle row of Figure~\ref{fig:nasbench}, containing the Gaussian Process based \gls{BO} experiments, shows that \our achieves a massive speed-up compared to both the adjacency and path encodings, the two dominant types of encoding in \gls{NAS} \cite{white2020study}. 
Perhaps owing to the high-dimensional, one-hot-encoded nature of the two alternative encoding, \gls{GP}-\gls{BO} struggles to learn meaningful relations and performs no more competitively than \gls{RS} or \gls{LS}, at least up to the query budget we consider 
(e.g. for a simple, $N$-node DAG, the adjacency encoding has $\frac{N(N-1)}{2} + N$ dimensions whereas the path encoding, without truncation, has  $\sum_{j=1}^N k^j$ dimensions \cite{white2020study}. In contrast, \our encoding only has $k$ dimensions corresponding to the $k$ operation choices).
On the other hand, by rigorously compressing the search space, for which non-parametric models such as \gls{GP}s are better suited, \our enables the full power of \gls{GP}-\gls{BO} to be utilised.

We show the results on \gls{NEP}-\gls{BO} in the bottom row of Figure~\ref{fig:nasbench}. The findings are consistent with those on \gls{GP}-\gls{BO} discussed above although Path-\gls{BO} performs much stronger. This is unsurprising, as \gls{NEP}-\gls{BO} is originally proposed to be used with the path encoding \cite{white2019bananas}, and that the hyperparameters we used are tuned for this purpose. Nonetheless, on \gls{NAS-Bench-201} search spaces, \our encoding delivers significant speedup, whereas on the more difficult \gls{NAS-Bench-301} search space, it is also significantly better in terms of final performance.

\paragraph{Comparison with SOTA}
We also compare our \our-based \gls{GP}-\gls{BO} with a wide range of competitive baselines, including reinforcement learning (\textsc{rl} \cite{ZophLe17_NAS}), regularised evolution (\textsc{re} \cite{real2019regularized}), Sequential Model-based Algorithm Configuration (\textsc{smac} \cite{hutter2011sequential}), Tree-structured Parzen Estimator (\textsc{tpe} \cite{bergstra2011algorithms}), graph convolutional network based \gls{BO} (\textsc{gcn}\gls{BO} \cite{shi2020multiobjective}), \gls{NEP}-\gls{BO} with path encodings (\textsc{bananas} \cite{white2019bananas}) and Weisfeiler--Lehman kernel-based \gls{BO} (\textsc{nas-bowl} \cite{wan2021interpretable}). Similarly, we compare \our-\gls{DNAS} with a number of existing \gls{DNAS} algorithms in Table \ref{tab:nasbench_diff}.
All additional details about the experimental setup can be found in App. \ref{app:experimental_setup}.

\begin{table}[!t]
    \centering
    \caption{Comparison of one-shot \gls{NAS} methods on \gls{NAS-Bench-201}. To reflect real-world applications, we search on \gls{CIFAR-10} and transfer the search result to the other datasets. }
\begin{footnotesize}
\resizebox{\columnwidth}{!}{%
\begin{tabular}{@{}llccc@{}}
\toprule
Benchmark & Search & \multicolumn{3}{c}{\gls{NAS-Bench-201}}  \\
Dataset & epoch & \gls{CIFAR-10} & \gls{CIFAR-100} & ImageNet16 \\
\midrule
\emph{Optimal} & - & $5.70$ & $26.50$ & $52.70$\\
\midrule
RSPS$^*$ \cite{li2020random} & $50$ & $12.34_{\pm 1.7}$ & $41.67_{\pm 4.3}$& $68.86_{\pm 3.9}$  \\
DARTS$^{\dagger}$ \cite{Liu2019_DARTS} & $50$ & $45.70_{\pm 0.0}$ & $84.39_{\pm 0.0}$ & $83.68_{\pm 0.0}$\\
SETN$^*$ \cite{dong2019one} & $50$ & $12.36_{\pm 0.0}$ & $41.95_{\pm 0.2}$ & $67.48_{\pm 0.2}$ \\
ENAS$^{\dagger} $ \cite{Pham2018_ENAS}   & $50$ & $45.70_{\pm 0.0}$ & $84.97_{\pm 0.0}$ & $83.68_{\pm 0.0}$ \\
GAEA-DARTS$^*$ \cite{li2020geometry} & $25$ & $8.36_{\pm 2.6}$ &  $31.61_{\pm 4.5}$ & $58.41_{\pm 4.2}$ \\
\midrule
\textbf{\our-\gls{DNAS}} & $20$ & $7.75_{\pm 1.2}$ & $31.33_{\pm 1.7}$ & $58.00_{\pm 2.9}$ \\
\bottomrule
\multicolumn{5}{l}{$^*$: Results taken from \cite{li2020geometry}; $^{\dagger}$: Results taken from \cite{dong2019NASbench201}.} \\
    \end{tabular}
    }
\end{footnotesize}
    \label{tab:nasbench_diff}
\end{table}

From Table \ref{tab:nasbench}, it is evident that \our-\gls{BO}  clearly outperforms competing methods, with the only \textsc{bananas} and \textsc{nas-bowl} being relatively close competitors in terms of final validation error.
For \textsc{nas-bowl}, we hypothesise that this could be because it relies on local mutation in the architecture space for candidate architecture generation (at each \gls{BO} iteration, the algorithm generates a pool of candidates from which \gls{BO} recommends the next evaluation), which might be overly greedy and exploitative and miss the better solution especially on larger and hence more complicated search spaces. For \textsc{bananas}, the performance gain mainly comes from the \gls{NEP} (which is orthogonal to \our), as we show previously, combining \our with \gls{NEP} also yields further improvements. With respect to Table \ref{tab:nasbench_diff}, despite of the much simpler formulation, \our-\gls{DNAS} performs competitively, outperforming all baseline methods significantly with the exception of \textsc{gaea}-\gls{DARTS} where they perform on par (note that we consider \textsc{gaea}-\gls{DARTS} with bi-level formulation since our implementation of \our-\gls{DNAS} is also bi-level. While the single level version of \textsc{gaea}-\gls{DARTS} outperforms the bi-level counterpart, this modification is orthogonal to our contribution, and hence could potentially be similarly exploited by \our-\gls{DNAS} for further performance gains). Furthermore, as we show the learning curves in Fig \ref{fig:nasbench_diff} in App. \ref{app:more_dnas}, \our-\gls{DNAS} almost instantly converges to a location of good solutions, allowing us to shorten the search time -- this is possible because \our encoding compresses and smooths the search space into a comparatively low-dimensional vector, allowing more effective gradient-based optimisation. Furthermore, in contrast to the previously mentioned performance collapse observed in some \gls{DNAS} methods \cite{dong2019NASbench201}, the performance of \our-\gls{DNAS} does not drop as optimisation proceeds.

\subsection{Open Domain Experiments}

For these experiments, we choose the popular \textsc{nasnet}-style search space \cite{Zoph2018_NASNet},
In our experiments, we use the setup of \textsc{lanas}  \cite{wang2019sample, wang2020learning} and due to the much higher computational cost, we only experiment with \our \gls{GP}-\gls{BO}. 
It is worth noting that due to the distributional nature of the \our encoding, differing from all previous \gls{NAS} methods, we do \emph{not} search for a single cell, but we instead search for an \emph{encoding}: at training and evaluation time, we randomly draw concrete realisations of architecture \emph{at each layer} by sampling the encoding according to the decoding scheme described in Section \ref{subsec:overview}. 
To highlight the sample efficiency of our method, during search we only sample and train $50$ architectures for $30$ epochs each. At the end of the search phase, we re-train the architecture corresponding to the best encoding seen to evaluate its performance.

\begin{table}[!t]
    \centering
    \caption{Performance on \gls{CIFAR-10} in the \textsc{nasnet}-style search space. \our-\gls{BO} experiment is conducted on $4\times$ NVIDIA Tesla V100 \textsc{gpu}s using $0.6$ wall-clock days.}
\begin{footnotesize}
\begin{tabular}{@{}lccc@{}}
\toprule
Algorithm & Val. Err &\#Params(M) & Method \\
\midrule
Random-WS \cite{xie2019exploring} & $2.85_{\pm0.08}$ & $4.3$ & RS \\
NASNet-A \cite{Zoph2018_NASNet} &$2.65$ & $3.3$ & RL \\
LaNAS \cite{wang2019sample} & $2.53_{\pm0.05}$ & $3.2$ & MCTS \\
DARTS \cite{Liu2019_DARTS}& $2.76_{\pm0.09}$  &$3.3$& GD\\
DARTS+$^{\dagger}$ \cite{liang2019darts+} & $2.37_{\pm 0.13}$ & $4.3$ & GD\\
P-DARTS \cite{chen2019progressive} & $2.50$ & $3.4$ & GD \\
DropNAS \cite{hong2020dropnas} & $2.58_{\pm 0.14}$ & $4.1$ & GD \\
DropNAS$^{\dagger}$ \cite{hong2020dropnas} & $1.88$ & $4.1$ & GD \\
 \midrule
BANANAS \cite{white2019bananas} & $2.64$ & - & BO\\
BOGCN \cite{shi2020multiobjective} & $2.61$ & $3.5$ & BO \\
NAS-BOWL \cite{wan2021interpretable} & $2.61_{\pm0.08}$ & $3.7$ & BO \\
\midrule
\textbf{\our-\gls{BO}} (Mixup) & $2.63$ & $3.5$ & BO \\
\textbf{\our-\gls{BO}} (CutMix) & $2.41$ & $3.5$ & BO \\
\textbf{\our-\gls{BO}+} & $1.86$ & $3.5$ & BO\\
\bottomrule
\multicolumn{4}{l}{$^{\dagger}$: Training protocol comparable to \our-\gls{BO}+} \\
\multicolumn{4}{l}{MCTS: Monte Carlo tree search; GD: Gradient descent.}
\end{tabular}
\end{footnotesize}
\label{tab:cifar10}
\end{table}

We report our results searched and evaluated on the \gls{CIFAR-10} dataset in Table \ref{tab:cifar10}, where we include two variants of \our-\gls{BO}, which only differ in the final evaluation: \our-\gls{BO}+ uses a range of additional enhancement techniques including CutMix \cite{yun2019cutmix}, Auxiliary Towers, DropPath \cite{larsson2016fractalnet}, AutoAugment \cite{cubuk2019autoaugment} and trains for an extended $1,500$ epochs; \our-\gls{BO} does not use AutoAugment and trains for the standard $600$ epochs and we report \our-\gls{BO} using two regularisation schemes.
Owing to both the shorter search budget and the smaller number of epochs per search (c.f. \textsc{bananas} \cite{white2019bananas}, which trains $100$ architectures for $50$ epochs ($\sim 12$ \textsc{gpu} days), and \textsc{lanas} \cite{wang2019sample} which trains $800$ architectures ($\geq 150$ \textsc{gpu} days)), \our-\gls{BO} is much more efficient than most other query-based \gls{NAS} methods.

\paragraph{Transferring to \gls{CIFAR-100}} We then transfer the best encoding searched on \gls{CIFAR-10} to the \gls{CIFAR-100} data-set, and we present the results against a number of competitive baselines in Table \ref{tab:cifar100} (note that some methods use advanced evaluation techniques similar to those used in \our-\gls{BO}+, and they are marked correspondingly). Remarkably, the performance obtained by us ($16.33 \%$ and $13.76\%$ for \our-\gls{BO} and \our-\gls{BO}+, respectively), outperforms some methods that are trained and searched on \gls{CIFAR-100} directly, empirically demonstrating the robustness of the \our encoding across various datasets.

\begin{table}[!t]
    \centering
    \caption{Performance on \gls{CIFAR-100} in the \textsc{nasnet}-style search space. The \our-\gls{BO} result is transferred from \gls{CIFAR-10}.}
\begin{footnotesize}
\begin{tabular}{@{}lccc@{}}
\toprule
Algorithm & Val. Err &\#Params(M) & Method \\
\midrule
DARTS \cite{Liu2019_DARTS} & $17.76$ & $3.3$ & GD \\
DARTS+$^\dagger$ \cite{liang2019darts+} & $14.87$ & $3.9$ & GD \\
P-DARTS$^*$ \cite{chen2019progressive} & $16.55$ & $3.4$ & GD \\ 
DropNAS \cite{hong2020dropnas} & $16.39$ & $4.4$ & GD\\
DropNAS+$^\dagger$ \cite{hong2020dropnas} & $14.10$ & $4.4$ & GD\\
 \midrule
\textbf{\our-\gls{BO}}$^*$ (CutMix) & $16.33$ & $3.5$ & BO \\
\textbf{\our-\gls{BO}}+$^*$ & $13.76$ & $3.5$ & BO \\
\bottomrule
\multicolumn{4}{l}{*: Transferred from the \gls{CIFAR-10} search.} \\
\multicolumn{4}{l}{$^\dagger$: Training protocol comparable to \our-\gls{BO}+.}
    \end{tabular}
\end{footnotesize}
    \label{tab:cifar100}
\end{table}

\section{Conclusion}

In this work, we have shown how \our, an approximate solution to the \gls{NAS} problem by effective compression of the search space, enables a speed-up in search with little to no deterioration in performance. 
Through combining with various search strategies, we show the versatility and effectiveness of our method on \gls{NAS} benchmarks.
We also perform \textit{open domain} experiments \gls{CIFAR-10}, showing competitive performances compared to the state-of-the-art and successfully transfer the encoding (and thus the architecture) to \gls{CIFAR-100}.
We believe the room for future works is ample: on one side, the operation distribution is a human-interpretable representation to communicate the \textit{gist} of the architecture. On the other hand, by not restricting the algorithm to a single architecture, \our can naturally generate diverse ensembles of related and high performing architectures. Ensembles have shown to provide benefits such as robustness~\cite{kariyappa2019improving} and future work should investigate this direction.
Furthermore, we have currently only validated the effectiveness on computer vision tasks in popular cell-based \gls{NAS} search spaces, and it would be similarly interesting to verify both alternative search spaces \cite{howard2019searching} and/or tasks \cite{klyuchnikov2020bench, mehrotra2020bench}. 
Finally, as opposed to most existing search strategies, \our is orthogonal to a large number of different strategies and thus could be effectively combined with ever-improving new strategies for even further improvements.

{\small
\bibliographystyle{ieee_fullname}
\bibliography{egbib}
}

\appendix

\begin{figure*}[t]
    \centering
    \begin{subfigure}{1\linewidth}
        \includegraphics[trim=1cm 7.5cm 0.5cm  1cm, clip, width=1.0\linewidth]{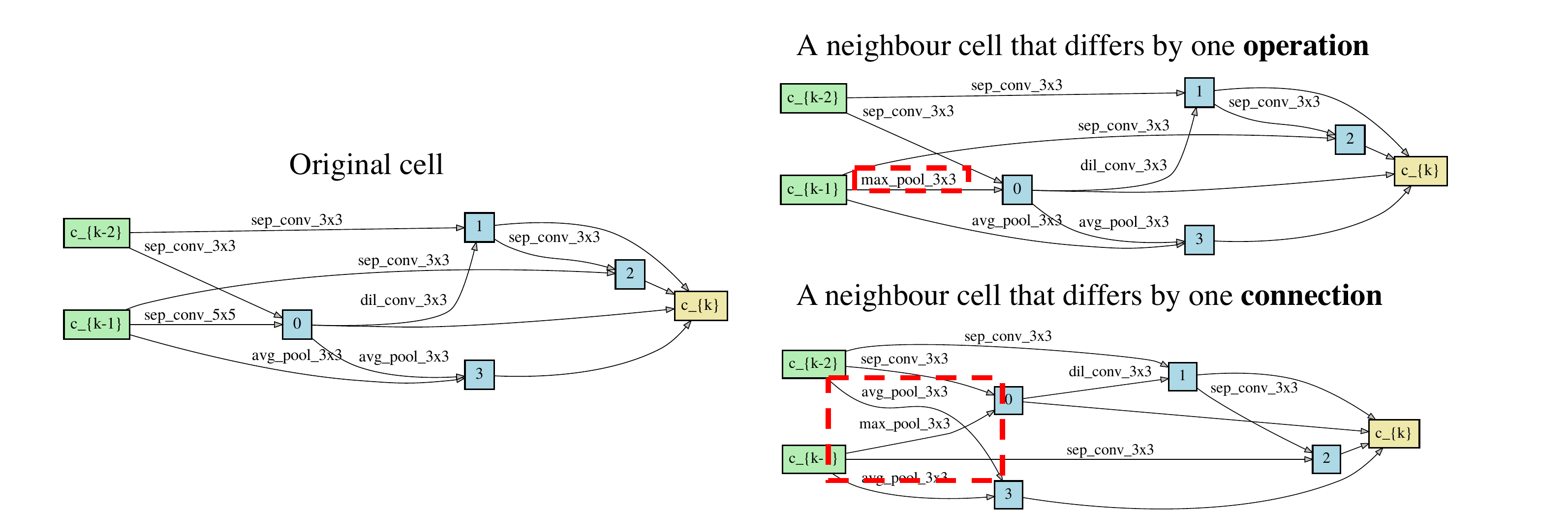}
    \end{subfigure}

    \caption{Examples of neighbouring cells in the architecture space. The first neighbour differs by one operation (circled in red) and the second neighbour differs by one connection (note the average pooling block is connected to input \texttt{c\_{\{k-2\}}} instead of node 0 in the original cell.
    }    
    \label{fig:neighbour_arch}
    \vspace{-4mm}
\end{figure*}

\newpage
\section*{Appendices}

\section{Experimental Setup}
\label{app:experimental_setup}
In this section we outline the detailed implementation setup for both our method and the baselines we implemented, on both the benchmark and open-domain tasks considered in this paper.

\subsection{Benchmark Tasks}
\paragraph{Data} We experiment on \gls{NAS-Bench-201} and \gls{NAS-Bench-301}. \gls{NAS-Bench-201} has a search space in the form of acyclic directed graph with 4 nodes and 6 edges, with each edge corresponding to a possible operation from \texttt{conv1$\times$1, conv$\times$3, skip-connect, maxpool and zeroize} (\texttt{zeroize} suggests the ending node will regard any input information as zero). The data-set contains 15,625 neural architectures, evaluated exhaustively on 3 image datasets, namely \gls{CIFAR-10}, \gls{CIFAR-100} and ImageNet16 (the downsampled ImageNet to $16 \times 16$). In addition to standard metrics such as validation and test accuracy/losses, we may also access additional information such as the complete training curve, time spent in training and FLOPs. The dataset and API can be downloaded from \url{https://github.com/D-X-Y/NAS-Bench-201}.

\gls{NAS-Bench-301} \cite{siems2020bench} differs from the previous tabular benchmark in that it contains a much larger search space (the search space is identical to the DARTS search space, which contains more than $10^{18}$ unique architectures). Each architecture contains 2 \emph{cells}, with each cell again being a acyclic directed graph similar to \gls{NAS-Bench-201}, but with 8 nodes and 14 possible edges. Each edge corresponds to one of a wider range of possible operations, namely separable convolution 3$\times$3, 5$\times$5; dilated convolution 3$\times$3, 5$\times$5; max pooling 3$\times$3, average pooling 3$\times$3, skip connection. Due to the prohibitively large search space it is not possible to exhaustively train each unique neural architecture contained in it. Instead, a large number of query-based \gls{NAS} methods, ranging from the simplest random search to more complicated methods such as \textsc{bananas} \cite{white2019bananas}, are run on this search space to give a good overall coverage by training a large number of architectures. With this large corpus of trained architectures and their corresponding accuracy, an ensemble of \emph{surrogate} models, including an ensemble of models including LGBoost\cite{ke2017lightgbm}, XGBoost \cite{chen2016xgboost} and graph isomorphic networks (GIN) \cite{xu2018powerful}, are fit so that given an unknown architecture $G$, we might use the \emph{predicted accuracy} from this surrogate model as a much cheaper proxy to the true, expensive accuracy that might only be obtained from training the architecture from scratch. Similar to \gls{NAS-Bench-201}, \gls{NAS-Bench-301} similarly provides the additional diagnostic information as listed above. The dataset and the API may be downloaded from \url{https://github.com/automl/nasbench301}.

\paragraph{Algorithm Setup} Unless otherwise stated, each method is repeated with 10 random starts and the mean/standard error are reported and where applicable, each method is also initialised with 10 random samples/evaluations at the beginning of the routines.

\begin{itemize}
    \item \textbf{Random Search (\gls{RS})} Both the original \gls{RS} and the unbiased \our-\gls{RS} are hyperparameter free, but due to the presence of bias away from the uniform distribution in \our-biased\gls{RS}, there exists some additional hyperparameters. In our experiments, we set the maximum Dirichlet scale $\beta_{\max}$ (explained in Section \ref{subsec:applyinganasod}) to be $2$ (i.e. the $\beta$ at the last random search iteration $t=T$), and we use a linear annealing schedule such that the $\beta$ at iteration $t$ is simply $t\frac{\beta_{\max}}{T}$.
    \item \textbf{Local Search (\gls{LS})} For \gls{LS}, we mainly adapt the implementation available in the repository associated with \cite{white2020study} (available at \url{https://github.com/realityengines/local_search} and \url{https://github.com/naszilla/nas-encodings}). For the baseline \gls{LS}, at each iteration the best performing architecture up to the current iteration $t$ ($\alpha^*_t$) is identified; From this, the algorithm exhaustively enumerates and queries its \emph{neighbours}, defined in this context as the other architectures $\alpha$ such that each of them only differs from $\alpha^*_t$ by edit distance of 1 (i.e. only different by 1 operation block or wiring. See Figure \ref{fig:neighbour_arch} for examples of neighbour architectures). The algorithm terminates (and restarts again if there is still unused budget) when it cannot find neighbour(s) to improve the best performance, suggesting that a (local) optimum has been reached.
    
    In \our-\gls{LS}, we use the same procedure but conduct the search in the \our encoding space instead of the architecture space at the beginning. Specifically, instead of identifying the best performing \emph{architecture}, we identify the best performing \emph{encoding} $\Tilde{\mathbf{p}}^*_t$ up to the current iteration $t$ and search its neighbours as described above. In this case, since each valid \our encoding is a vertex on the regular grid in the $k$-simplex, the neighbours are simply the adjacent vertices on the grid (for example, for a cell with $N=5$ and $k=3$, the neighbours of encoding $[1, 1, 3]$ are $[0, 1, 4], [1,0,4], [0,2,3], [2,0,3], [1,2,2]$ and $[2,1,2]$).
    
    \item \textbf{\our-\gls{BO}} We use a standard Gaussian Process (\gls{GP}) as the surrogate for \our-\gls{BO} with M\'atern 5/2 kernel. To improve the goodness of fit of \gls{GP}, we always normalise the input (in our case, the \our encodings) into standard hyperrectangular boxes $[0, 1]^k$. For the targets (in our case, the validation error), we find that using a log-transform often leads to better estimation of the predictive uncertainty, similar to the finding in \cite{wan2021interpretable}. We then use a normalisation wrapping around the log-predictive uncertainty by de-meaning the data and dividing by the data standard deviation such that the \gls{GP} targets are approximately within the range of $\mathcal{N}(0, 1)$.
    
    For the \gls{GP} hyperparameters, we constrain the lengthscales to be within $[0.01, 0.5]$, the outputscale to be in $[0.5, 5]$ and the noise variance to be within $[10^{-6}. 10^{-1}]$ and by default we do not enable the automatic relevance determination (ARD). The exact hyperparameters of the \gls{GP} are obtained by optimising the \gls{GP} log-marginal likelihood and the \gls{GP} model is implemented using the gpytorch \cite{gardner2018gpytorch} package which feature automatic differentiation.
    
    For the implementation of the \gls{BO} routine, we use the expected improvement criterion as the acquisition function that needs to be optimised at each iteration. To optimise the acquisition function to obtain the recommendation from the \gls{BO} $\arg \max a(\Tilde{\mathbf{p}})$, in this paper we use the sample-based optimisation by first randomly generating $100k$ samples via randomly sampling the \emph{candidate generating distribution} (described in Algorithm \ref{alg:bo} in the main text). However, we also explore a gradient-based optimisation approach in to the acquisition function, whose update may be written as:
    \begin{equation*}
        \Tilde{\mathbf{p}}' \leftarrow \Tilde{\mathbf{p}} - \eta \frac{\partial a(\Tilde{\mathbf{p}})}{\partial \Tilde{\mathbf{p}}}
    \end{equation*}
    \begin{equation}
        \Tilde{\mathbf{p}} \leftarrow \frac{\Tilde{\mathbf{p}}' }{\sum_i^k \Tilde{p_i} }
        \label{eq:simplexconstraint}
    \end{equation}
    where we conduct mirror descent by combining the gradient-based update (we use the simple gradient descent for simplicity but the optimisation is compatible with gradient-based optimisers such as Adam; here $\eta$ is the generic learning rate for gradient-based methods) with the normalisation to enforce the simplex constraint. We find that both acquisition optimisation methods to be broadly comparable, but it is trivial in concept and implementation to extend the sample-based optimisation technique into batch setting where at each \gls{BO} iteration, multiple recommendations are produced.
    
    \item \textbf{Other baselines} For reinforcement learning (\textsc{rl}), regularised evolution (\textsc{re}), Sequential Model-based Algorithm Configuration (\textsc{smac}), Tree-structured Parzen Estimator (\textsc{tpe}), we use the code available at \url{https://github.com/automl/nas_benchmarks/} for the \gls{NAS-Bench-201} search space and adapt the codes for the \gls{NAS-Bench-301} repository. For \textsc{bananas} \cite{white2019bananas}, we use the official implementation available at \url{https://github.com/naszilla/bananas}; for \textsc{gcn}-based \gls{BO}, we use the implementation available at NASLib \cite{ruchte2020naslib} \url{https://github.com/automl/NASLib}; for \textsc{nas-bowl} \cite{wan2021interpretable}, we use the official implementation at \url{https://github.com/xingchenwan/nasbowl}. Unless otherwise specified, we use the default settings and hyperparameters in these methods. Since \gls{NAS-Bench-301} is to be used as a cheaper proxy to the \gls{DARTS} search space, for the baseline methods, we use the default settings for \gls{DARTS} search as the settings for the \gls{NAS-Bench-301} experiments. 
\end{itemize}

\subsection{Open-domain Tasks}
\label{subsec:opendomain}

For the open-domain tasks we use the popular \textsc{nasnet}-style \cite{Zoph2018_NASNet} search space. Since \gls{NAS-Bench-301} benchmark already includes the typical \gls{DARTS} search space (i.e. \textsc{nasnet}-style search space with 8 nodes and 14 possible edges), here we follow previous works \cite{wang2020learning, wang2019sample} by opting for an enlarged search space with 10 nodes and 20 possible edges, containing order-of-magnitudes more unique architectures even than the standard \gls{DARTS} space -- this makes the search of a good performing cell within a small number of queries even more challenging. For the implementation details, we largely follow \cite{wang2020learning, wang2019sample, Liu2019_DARTS} but with two major exceptions: first, as a demonstration of our sample efficiency, we both train much fewer \emph{architectures} during search time, and train fewer \emph{epochs per architecture} compared to previous works such as \cite{wan2021interpretable, white2019bananas, shi2020multiobjective}: during search time we use the \our-\gls{BO} procedure described above. Second, while \cite{wang2019sample} restricts the candidate operations to $4$ perceived to be high-performing (max pool, separable convolution 3$\times$3, skip connection and dilated convolution 3$\times$3), we retain the full range of $8$ operation candidates for fair comparison with other previous works. For an encoding $\Tilde{\mathbf{p}}$, we sample 24 valid cells randomly to build the architecture (this is possible due to the fact that we may sample architectures from the \our encoding; previous work typically repeats the identical cell multiple times to build a network), and we train each sampled architecture for 30 epochs using SGD with momentum optimiser with cosine annealed learning rate schedule (learning rate at start $\eta_0 = 0.025$ which anneals to 0 at the 30th epoch; momentum $= 0.9$; weight decay = $3 \times 10^{-4}$) with a batch size of 96, depth of 24 and initial channel number of 36. We terminate the search after training 50 architectures as described, and select the encoding that leads to the best validation error for \emph{evaluation}. To utilise the parallel computing facilities, we run \our-\gls{BO} with batch size of 4 (at each \gls{BO} iteration, 4 encodings are generated and recommended for training); it is worth noting that further parallelisation is possible to reduce the wall clock time. For example, it is possible to trivially reduce the wall-clock time to 0.3 days if 8 \textsc{gpu}s are used.

\begin{figure*}[t]
    \centering
    \begin{subfigure}{0.25\linewidth}
        \includegraphics[trim=0cm 0cm 0cm  0cm, clip, width=1.0\linewidth]{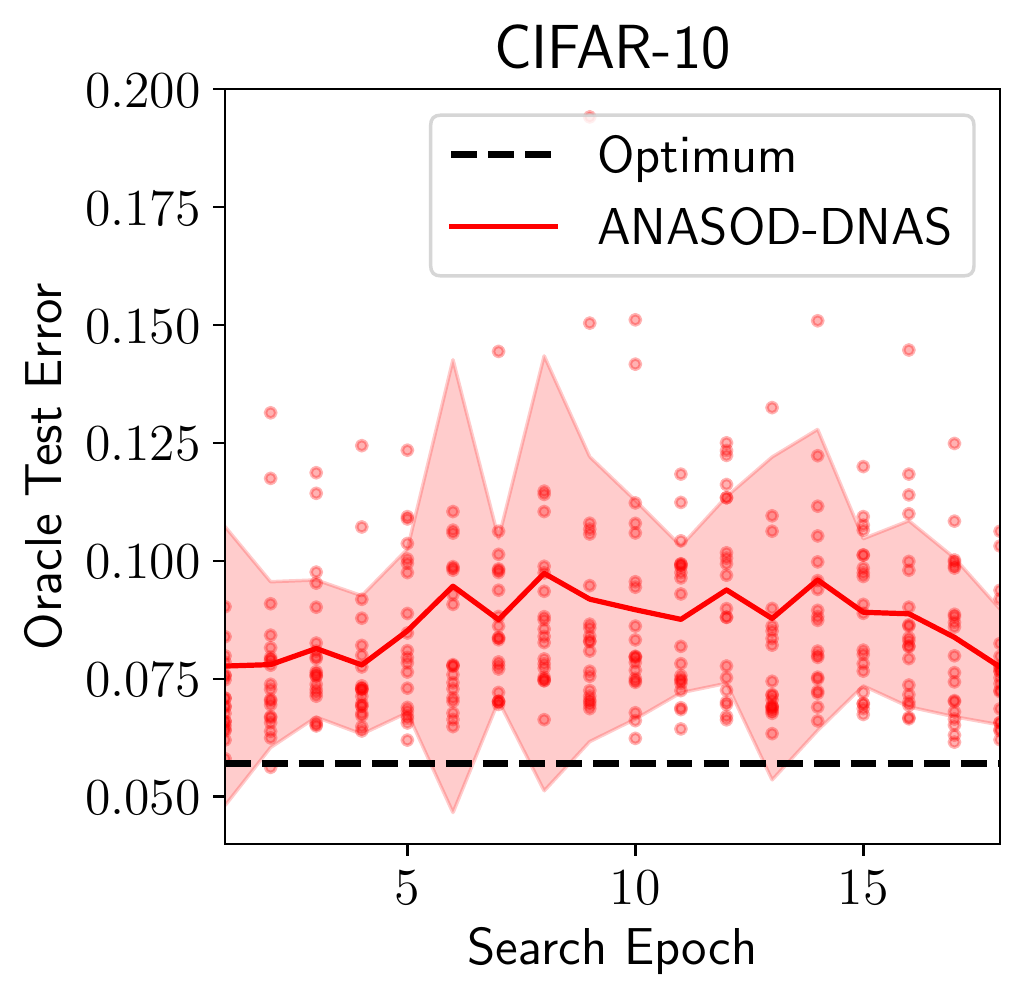}
    \end{subfigure}
    \begin{subfigure}{0.25\linewidth}
        \includegraphics[trim=0cm 0cm 0cm  0cm, clip, width=1.0\linewidth]{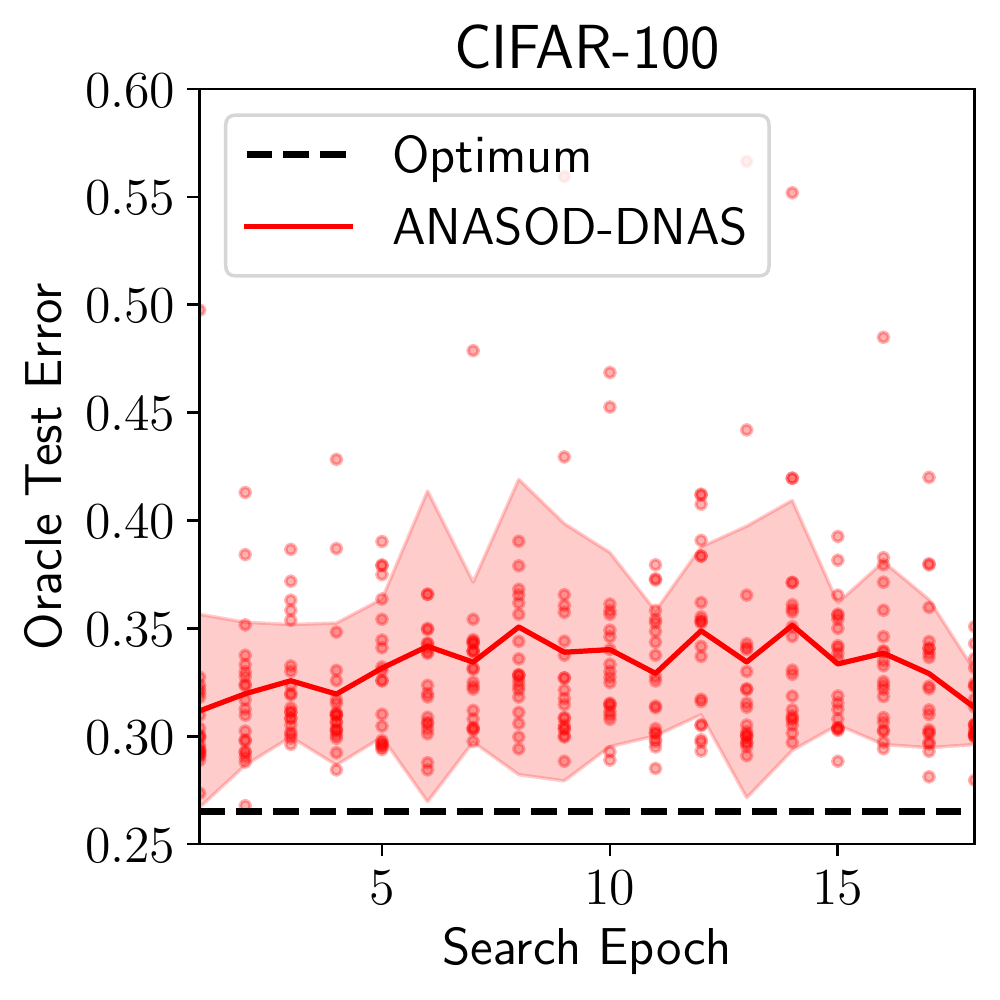}
    \end{subfigure}
        \begin{subfigure}{0.25\linewidth}
        \includegraphics[trim=0cm 0cm 0cm  0cm, clip, width=1.0\linewidth]{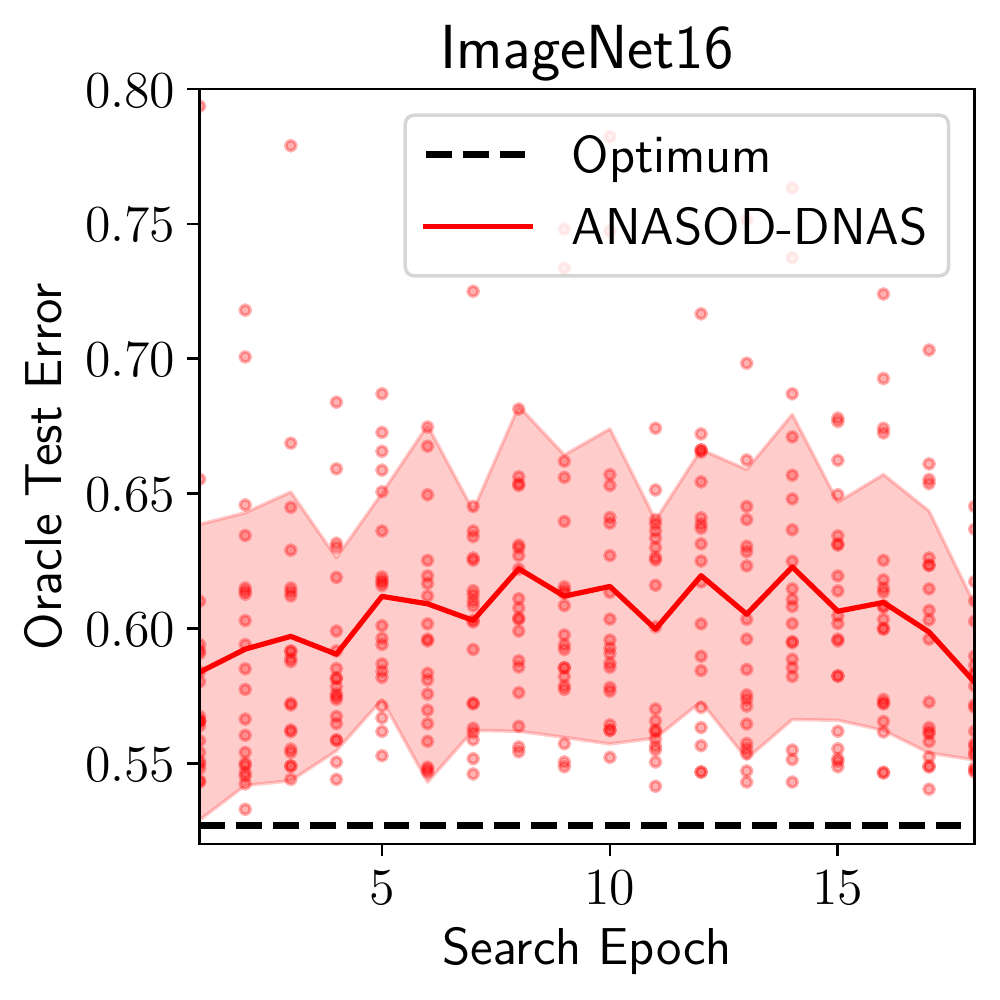}
    \end{subfigure}

    \caption{Learning curves of \our-\gls{DNAS} on \gls{NAS-Bench-201} datasets (oracle test error vs the number of learning epochs). We search on the \gls{CIFAR-10} training set and transfer the searched cell directly to the other two datasets. Lines and shades denote mean $\pm$ 1 standard deviation, and dots denote the data points every each trial. Note that unlike previous methods such as \gls{DARTS} that quickly collapse to all-skip-connection cells that perform poorly, \our-\gls{DNAS} converges quickly to good solutions and maintains high performance throughout the optimisation process.
    }    
    \label{fig:nasbench_diff}
\end{figure*}

We use two slightly different evaluation procedures and thus report two results, namely \our-\gls{BO} and \our-\gls{BO}+ in Tables \ref{tab:cifar10} and \ref{tab:cifar100} where we largely follow \cite{wang2019sample}. For the standard, \our-\gls{BO} result, we train the architecture induced by the encoding for 600 epochs, again with the SGD with momentum optimiser with initial learning rate set at $\eta_0 = 0.025$ (but the learning rate schedule is now annealed across the entire 600 epochs instead of the 30 epochs described above) and every other hyperparameter is kept consistent with the description above for search. During evaluation, we also include additional augmentation techniques such as CutMix \cite{yun2019cutmix}, Auxiliary Tower with probability 0.4 and drop path with probability 0.2. For \our-\gls{BO}+, we follow previous works such as \cite{liang2019darts+} to increase the training to 1,500 epochs with identical optimiser setting (with the learning rate annealed over 1,500 epochs), and we additionally use AutoAugment. We use the default train/validation partition sets of \gls{CIFAR-10} for both search and evaluation, and for the \gls{CIFAR-100} transfer learning experiment, we directly use the cell searched from \gls{CIFAR-10} for evaluation. The evaluation protocol on \gls{CIFAR-100} is otherwise identical to that on \gls{CIFAR-10} described above.

\section{Results on Local Search}
\label{app:local_search}

Local search has recently been shown to be an extremely powerful optimiser, achieving state-of-the-art performance especially in small search spaces \cite{white2020local}. However, a well-known theoretical and empirical drawback of \gls{LS} is its susceptibility of getting trapped locally \cite{hromkovivc2013algorithmics, aarts2003local}, especially when the search space is large and/or is rife with local optima (in the \gls{NAS} context, \gls{LS} is found to perform much worse in larger search spaces (e.g. \gls{DARTS} space) both in \cite{white2020local} and by us in Sec \ref{subsec:nasbench}). To leverage the \our encoding, we propose a simple modification (which we term \our-\gls{LS} in Algorithm \ref{alg:ls}) which performs \gls{LS} on the \our encoding space first and switches to the architecture space when the algorithm fails to improve (i.e. it reaches a local optimum). As discussed, \our encoding effectively abstracts the huge, original architecture space by compressing multiple unique architectures into a single encoding with fewer neighbours (a neighbour in encoding space is an adjacent vertex on the regular grid on the $k$-simplex), both of which allow \gls{LS} on the encoding space to make rapid progress on a smoothed objective function landscape. It is worth noting that for this acceleration to be possible, we have to both 1) be able to easily sample architectures \emph{from} the encoding and 2) have an encoding scheme that compresses and smooths the original architecture search space so that \gls{LS} is less likely to be stuck. To our knowledge, \our is the only one that meets both desiderata.

\begin{figure*}[t]
    \centering
    
         \begin{subfigure}{0.24\linewidth}
    \includegraphics[trim=0cm 0cm 0cm  0cm, clip, width=1.0\linewidth]{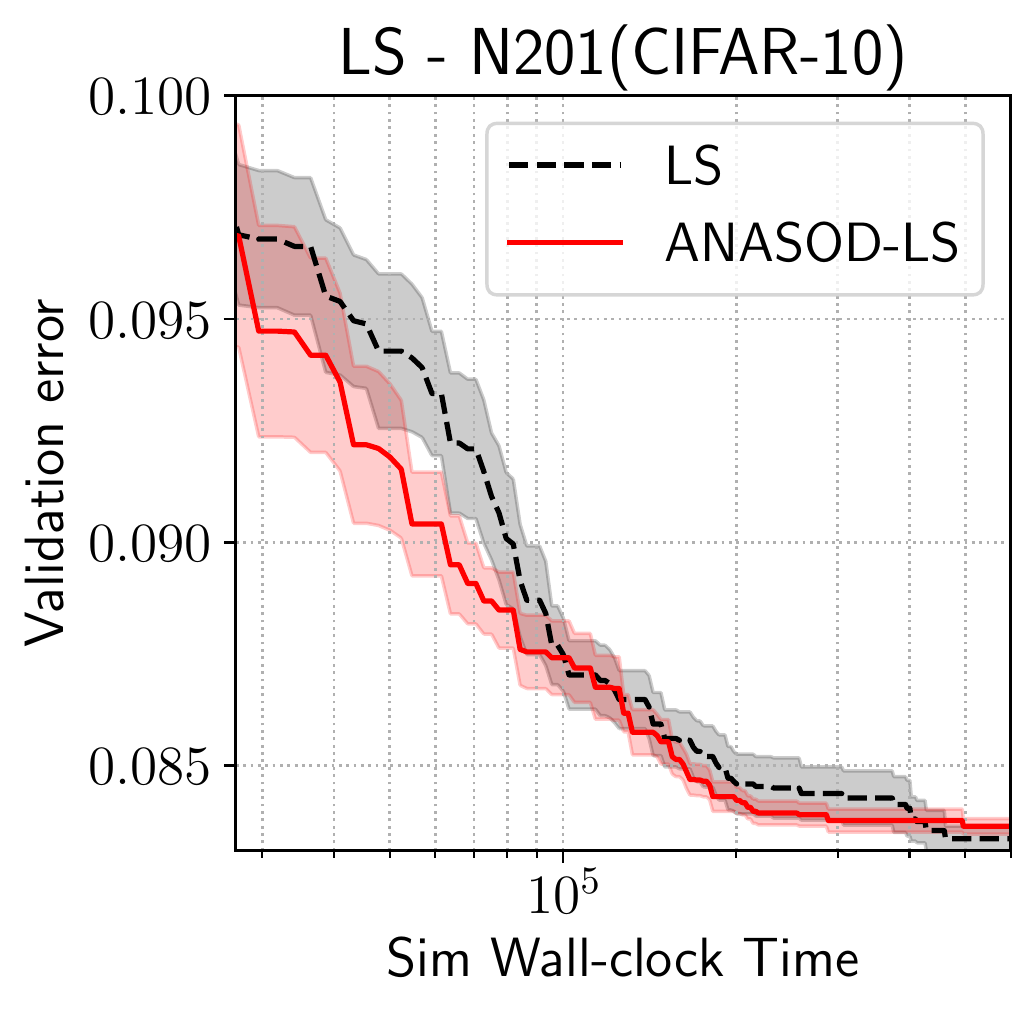}
    \end{subfigure}
    \begin{subfigure}{0.225\linewidth}
        \includegraphics[trim=0cm 0cm 0cm  0cm, clip, width=1.0\linewidth]{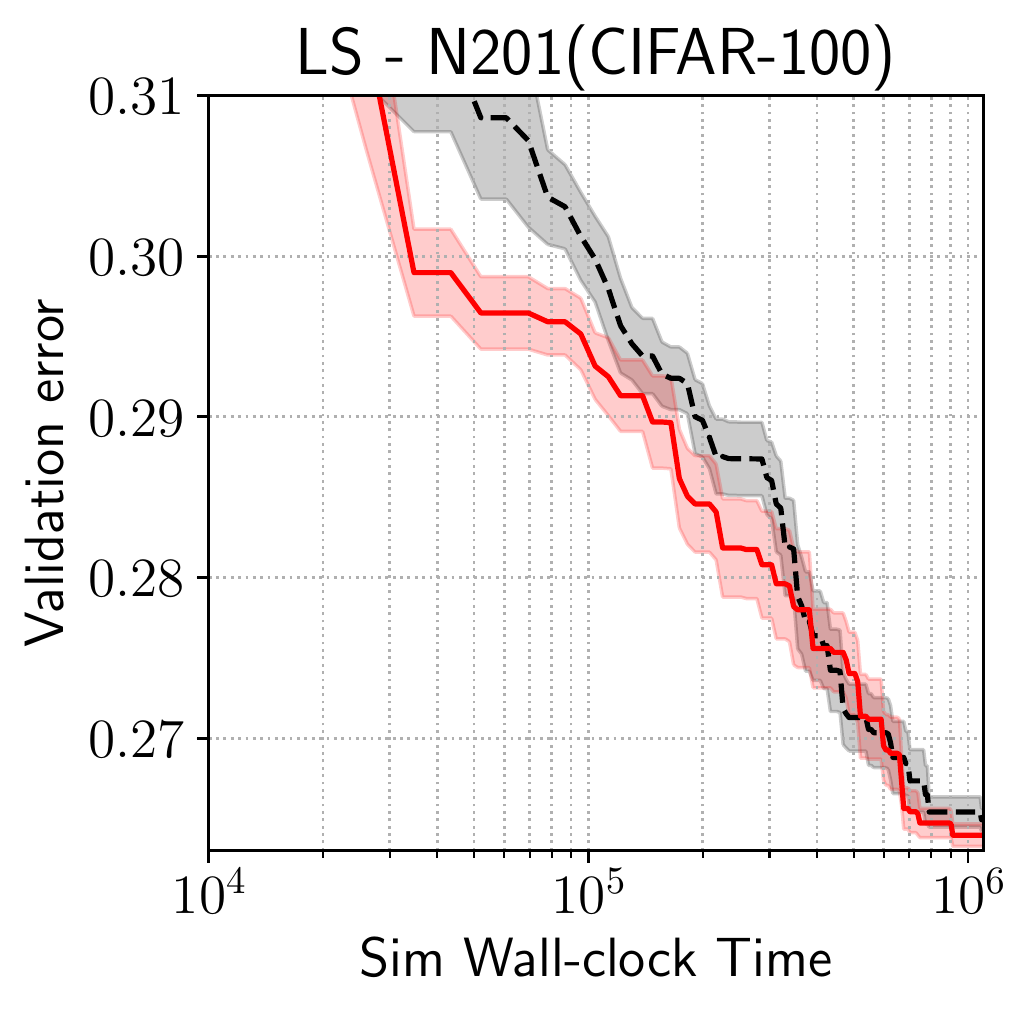}
    \end{subfigure}
         \begin{subfigure}{0.22\linewidth}
    \includegraphics[trim=0cm 0cm 0cm  0cm, clip, width=1.0\linewidth]{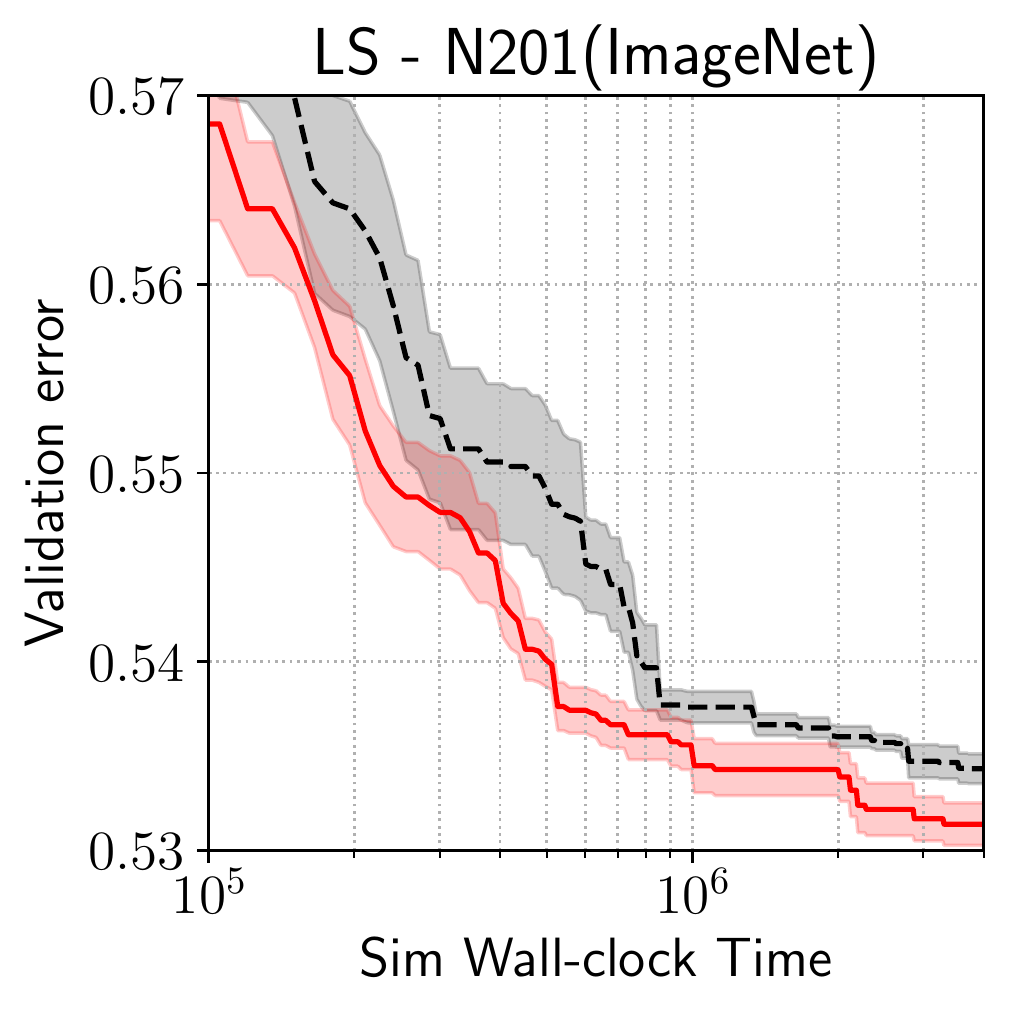}
    \end{subfigure}
    \begin{subfigure}{0.23\linewidth}
    \includegraphics[trim=0cm 0cm 0cm  0cm, clip, width=1.0\linewidth]{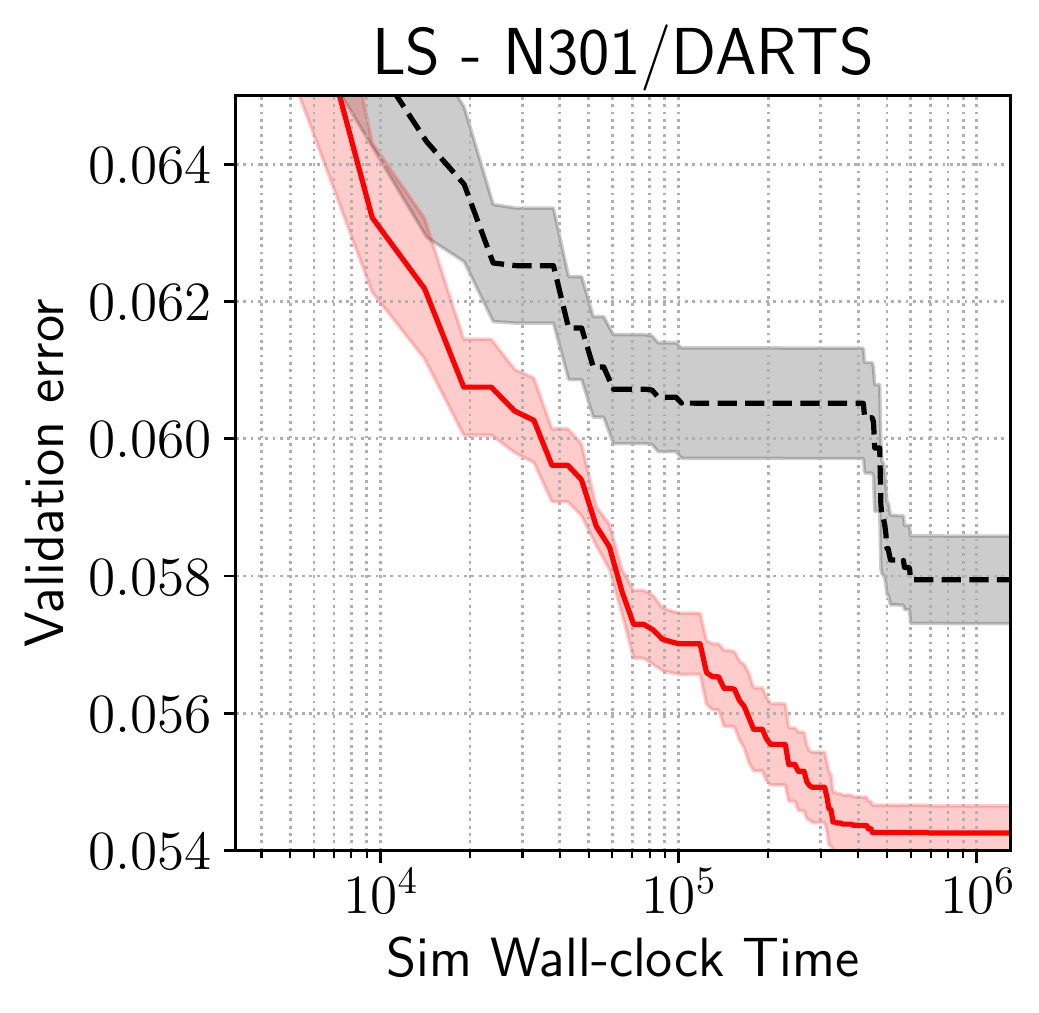}
    \end{subfigure}
    
    \caption{Performance of \gls{LS} vs \our-\gls{LS} on \gls{NAS-Bench-201} and \gls{NAS-Bench-301}. Lines and shades denote mean $\pm$ 1 standard error, across 10 different random trials.
    }    
    \label{fig:nasbench_local_search}
\end{figure*}

We conduct experiments of \our-\gls{LS} on \gls{NAS-Bench-201} and \gls{NAS-Bench-301} search spaces, and the results are shown in Table \ref{tab:nasbench_localsearch} and Fig \ref{fig:nasbench_local_search}. In line with the observation of \cite{white2020local}, we observe that the baseline \gls{LS} performs extremely competitively in the \gls{NAS-Bench-201} tasks but struggles in the larger \gls{NAS-Bench-301} search space and under-performs even random search---presumably due to the more prevalent local optima that prevents \gls{LS} from progressing further. 
However, \our-\gls{LS} restores the competitiveness in \gls{NAS-Bench-301}, and even for the \gls{NAS-Bench-201} tasks where the original \gls{LS} already performs strongly, the use of \our nonetheless improves its convergence speed in the initial stage of the optimisation. Generally, our experiments show that the margin of improvement increases as the task difficulty increases.
Indeed, in the ImageNet task \our improves both convergence speed and final performance; and in the CIFAR tasks \our does not improve the final performance further as the baseline is already very close to the ground-truth optimum.

\begin{table}[!t]
    \centering
    \caption{Comparison of original \gls{LS} with \our-\gls{LS}. We report mean $\pm$ 1 standard error of the validation error across 10 random trials. Numbers shown denote the best validation error seen after $\mathbf{300}$ architecture queries.}
\begin{footnotesize}
\resizebox{\columnwidth}{!}{%
\begin{tabular}{@{}lcccc@{}}
\toprule
Benchmark & \multicolumn{3}{c}{\gls{NAS-Bench-201}} & \gls{NAS-Bench-301} \\
Dataset & \gls{CIFAR-10} & \gls{CIFAR-100} & ImageNet16 & \gls{CIFAR-10} \\
\midrule
LS \cite{white2020local} & $\mathbf{8.34}_{\pm 0.03}$ & $26.41_{\pm 0.04}$ & $53.43_{\pm 0.08}$ & $5.79_{\pm 0.06}$\\
\textbf{\our-LS} & $8.36_{\pm 0.02}$ & $\mathbf{26.38}_{\pm 0.07}$ & $\mathbf{53.07}_{\pm 0.09}$ & $\mathbf{5.43}_{\pm 0.04}$\\
\bottomrule
    \end{tabular}
    }
\end{footnotesize}
    \label{tab:nasbench_localsearch}
\end{table}

\begin{algorithm}[!ht]
\begin{footnotesize}
	    \caption{Local search. Steps added or modified by \our-\gls{LS} are marked \textcolor{blue}{blue}.}
	    \label{alg:ls}
	\begin{algorithmic}[1]
		\STATE {\bfseries Input:} Search space $A$, objective function (default: validation error) $y$
		\STATE \textcolor{blue}{Sample an \our encoding at uniform $\Tilde{\mathbf{p}_1} \sim \mathrm{Dir}(1,...,1)$}
		\STATE \textcolor{blue}{Evaluate a single architecture $\alpha_1$ from the encoding $\Tilde{\mathbf{p}_1}$ to approximate the performance of the all architectures parameterised by $\Tilde{\mathbf{p}_1}$: $y(\Tilde{\mathbf{p}_1}) = \mathbb{E}_{\alpha \sim p(\alpha|\Tilde{\mathbf{p}})} \Big[y({\alpha})\Big] \approx y(\alpha_1), \alpha_1 \sim p(\alpha|\Tilde{\mathbf{p}})$; denote dummy variable $y(\Tilde{\mathbf{p}}_0) = \infty$.}
		\WHILE{\textcolor{blue}{$y(\Tilde{\mathbf{p}}_i) \leq y(\Tilde{\mathbf{p}}_{i-1})$}}
		\STATE \textcolor{blue}{Search and evaluate the neighbourhood on the \textbf{encoding space} $\Tilde{\mathbf{p}}_{i+1} \leftarrow \mathrm{SearchNeighbour} (\Tilde{\mathbf{p}}_{i})$.}
		\ENDWHILE
		\STATE \textcolor{blue}{/* \emph{Switch to arch space when \gls{LS} on encoding reaches an optimum} */}
		\IF{\textcolor{blue}{Using \our-\gls{LS}}}
		\STATE \textcolor{blue}{Sample an arch \emph{from the best encoding found }:  $\alpha_i \sim p(\alpha|\Tilde{\mathbf{p}}^*)$}
		\ELSE
		\STATE Sample an arch uniformly from the search space $A$.
		\ENDIF 
		\WHILE{$y(\Tilde{{\alpha}}_i) \leq y(\Tilde{{\alpha}}_{i-1})$}
		\STATE Search and evaluate the neighbourhood on the \textbf{architecture space} $\Tilde{\alpha}_{i+1} \leftarrow \mathrm{SearchNeighbour} (\Tilde{\alpha}_{i})$.
		\ENDWHILE
	\end{algorithmic}
\end{footnotesize}
\end{algorithm}

\section{Further results on ANASOD-DNAS}
\label{app:more_dnas}

\begin{figure*}[t]
    \centering
    \begin{subfigure}{1\linewidth}
        \includegraphics[trim=1cm 5cm 0.5cm  4.5cm, clip, width=1.0\linewidth]{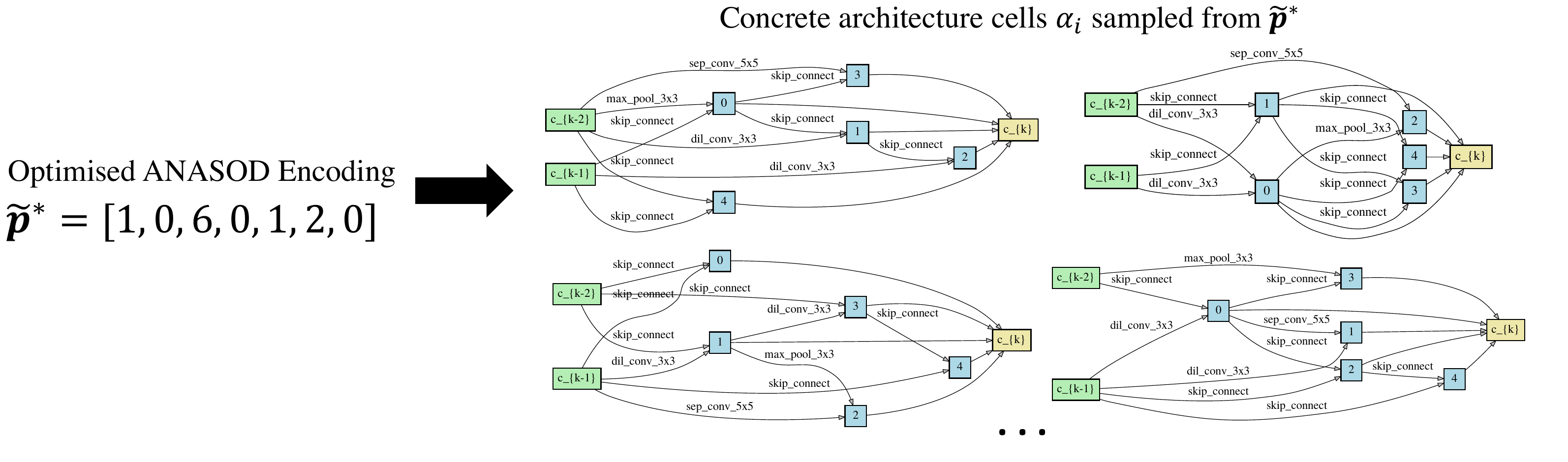}
    \end{subfigure}

    \caption{Best encoding found by \our-\gls{BO} in the open-domain task (i.e. 1 max pooling, 6 skip connections, 1 separable convolution $5\times5$ and 2 dilated convolution $3\times3$) and examples of concrete cells sampled from the optimised \our encoding. Note that, similar to  \cite{white2019bananas, wang2019sample, wan2021interpretable} who use the same cell, we sample the normal and reduction cells from the same encoding.
    }    
    \label{fig:best_encode}
    \vspace{-4mm}
\end{figure*}

We show the learning curves of \our-\gls{DNAS} in Figure \ref{fig:nasbench_diff}, where at each training epoch we sample and query the oracle test accuracy of an architecture from the encoding learnt at that point: \our-\gls{DNAS} almost \emph{instantly} finds and converges to a location of good solutions, allowing us to shorten the search time while still recovering good solutions -- this is possible because \our encoding massively compresses and smooths the search space into a comparatively low-dimensional vector, allowing gradient-based optimisers to converge efficiently. Furthermore, unlike methods like \gls{DARTS} that are shown to lack sufficient regularisation and quickly overfits to find cells dominated by skip-connections that hence perform poorly \cite{dong2019NASbench201}, the performance of \our-\gls{DNAS} does not drop as optimisation proceeds.

\paragraph{Implementation details} We mainly use the repository open-sourced by the authors of \cite{li2020geometry} available at \url{https://github.com/liamcli/gaea_release}. Implementing \our-\gls{DNAS} on the basis of \cite{li2020geometry} is extremely simple: instead of searching for the optimal categorical distribution parameters on each edge, we force all edges to share one single set of parameters (i.e. the \our encoding). We use the default hyperparameters settings as in \cite{li2020geometry}. 

\section{Optimal Encoding Searched on the Open-domain Task}
\label{app:optimal_encoding}
In this section we present the best encoding $\Tilde{\mathbf{p}}^*$ found by \our-\gls{BO} in the open-domain search space described in Section \ref{subsec:opendomain} (Figure \ref{fig:best_encode}). Unlike existing \gls{NAS} method which searches for an exact \emph{cell}, the optimal encoding maps to a distribution of architectures from which we sample concrete cells to build the neural network for evaluation, and some examples of which are also demonstrated in Figure \ref{fig:best_encode}. It is very interesting that \our-\gls{BO} produces light-weight cells with a large number of skip connections but nevertheless perform promisingly (i.e. the skip connections might indeed help in the performance, as opposed to being an artefact of overfitting). Furthermore, the cells obtained are very different from the optimised cells from most methods that are dominated by separable convolution $3\times3$; this suggests that the cells dominated by separable convolution $3\times3$ might be only one of the local minima in the loss landscape w.r.t the \emph{architecture} parameters; other possible optima on the possibly multi-modal landscape, while not often discovered by the existing methods, might perform equally competitively. We believe the characterisation and analysis of the loss landscape w.r.t architecture parameters , a heretofore under-explored area, could be an exciting future direction for further theoretical understanding of \gls{NAS}.

\end{document}